\documentclass[10pt,twocolumn,letterpaper]{article}

\usepackage{cvpr}
\usepackage{times}
\usepackage{epsfig}
\usepackage{graphicx}
\usepackage{amsmath}
\usepackage{booktabs}
\usepackage{amssymb}
\usepackage{url}

\usepackage{subcaption}
\usepackage{dblfloatfix}    % To enable figures at the bottom of page
\newcommand*{\affmark}[1][*]{\textsuperscript{#1}}
\usepackage[breaklinks=true,bookmarks=false]{hyperref}

\cvprfinalcopy % *** Uncomment this line for the final submission

\def\cvprPaperID{****} % *** Enter the CVPR Paper ID here

\begin{document}

%%%%%%%%% TITLE
\title{Heuristics2Annotate: Efficient Annotation of Large-Scale Marathon Dataset For Bounding Box Regression}

% \author{Pranjal Singh Rajput\\
% Computer Vision Lab\\ TU Delft
% % Delft University of Technology\\ The Netherlands\\
% % {\tt\small pranjal.manit@gmail.com}
% \and
% Supervisor: Yeshwanth Napolean\\
% Computer Vision Lab\\ TU Delft
% % Delft University of Technology\\ The Netherlands\\
% % {\tt\small secondauthor@i2.org}

% \and
% Supervisor: Jan van Gemert\\
% Computer Vision Lab\\ TU Delft
% % Delft University of Technology, The Netherlands\\
% % {\tt\small secondauthor@i2.org}\\
% % Computer Vision Lab\\
% % Delft University of Technology\\ The Netherlands\\
% }

\author{Pranjal Singh Rajput\affmark[1] \and Yeshwanth Napolean\affmark[1] \and Jan van Gemert\affmark[1]
\and \affmark[1]Computer Vision Lab,  Delft University of Technology\\
}

% \author{Pranjal Singh Rajput\qquad Yeshwanth Napolean \qquad Jan van Gemert \\ \\ Computer Vision Lab \\ Delft University of Technology, The Netherlands}
\maketitle
%\thispagestyle{empty}

%%%%%%%%% ABSTRACT
%   The ABSTRACT is to be in fully-justified italicized text, at the top
%   of the left-hand column, below the author and affiliation
%   information. Use the word ``Abstract'' as the title, in 12-point
%   Times, boldface type, centered relative to the column, initially
%   capitalized. The abstract is to be in 10-point, single-spaced type.
%   Leave two blank lines after the Abstract, then begin the main text.
%   Look at previous CVPR abstracts to get a feel for style and length.
\begin{abstract}
Annotating a large-scale in-the-wild person re-identification dataset especially of marathon runners is a challenging task. The variations in the scenarios such as camera viewpoints, resolution, occlusion, and illumination make the problem non-trivial. Manually annotating bounding boxes in such large-scale datasets is cost-inefficient. Additionally, due to crowdedness and occlusion in the videos, aligning the identity of runners across multiple disjoint cameras is a challenge. We collected a novel large-scale in-the-wild video dataset of marathon runners. The dataset consists of hours of recording of thousands of runners captured using 42 hand-held smartphone cameras and covering real-world scenarios. Due to the presence of crowdedness and occlusion in the videos, the annotation of runners becomes a challenging task. We propose a new scheme for tackling the challenges in the annotation of such large dataset. Our technique reduces the overall cost of annotation in terms of time as well as budget. We demonstrate performing fps analysis to reduce the effort and time of annotation. We investigate several annotation methods for efficiently generating tight bounding boxes. Our results prove that interpolating bounding boxes between keyframes is the most efficient method of bounding box generation amongst several other methods and is 3x times faster than the naive baseline method. We introduce a novel way of aligning the identity of runners in disjoint cameras. Our inter-camera alignment tool integrated with the state-of-the-art person re-id system proves to be sufficient and effective in the alignment of the runners across multiple cameras with non-overlapping views. Our proposed framework of annotation reduces the annotation cost of the dataset by a factor of 16x, also effectively aligning 93.64\% of the runners in the cross-camera setting.

    \end{abstract}

%%%%%%%%% BODY TEXT

\begin{figure}[ht]
\begin{center}
   \includegraphics[width=\linewidth]{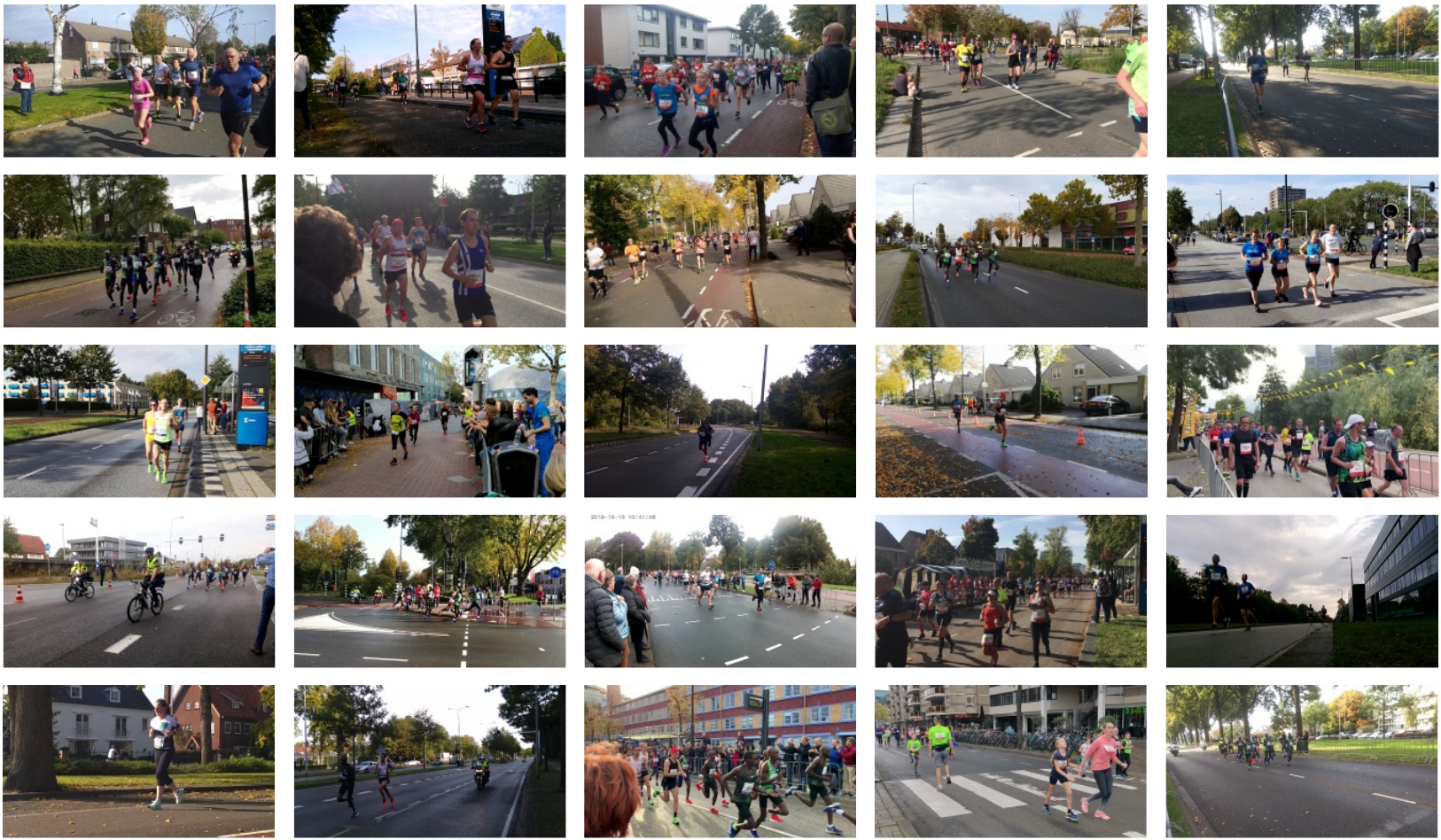}
\end{center}
   \caption{\textbf{Sample images of videos recorded from multiple disjoint cameras.} Different scenarios including variation in light intensity, occlusion, shadows, and different camera poses and angles make the task of runner detection non-trivial.}
\label{fig:imageGrid}
\end{figure}

\section{Introduction}
Marathon events are gaining popularity due to cognizance towards health awareness and motivation in improving psychological well being. The professional runners always aim in improving their performances. Nowadays, athletes record their running data for their personal health monitoring. Also, some event organizers record the race along with the individual athlete's statistics for their performance tracking. Sometimes runners are interested in retrieving their moments of appearance in different time frames during the event to monitor their performance during the race. Retrieving such images and videos for thousands of participants from numerous videos is a challenging task. Individually runners can be tracked using a personalized GPS tracker, but the videos and images contain multiple athletes at a time.

Searching a runner with just a single portrait, in thousands of videos with hundred hours of recordings from disjoint cameras is a laborious task. Every runner has a unique bibId, a unique number attached in front of its uniform at the chest. If the bibId is visible, then text recognition models can be used to identify and locate the athlete. We can also use computer vision-based person re-identification models to identify the runner if a clear portrait of the runner is available. But in the real scenarios, the athletes are partially or fully occluded, having similar clothing, different poses, variable illumination in images/videos making it challenging even for state-of-the-art computer vision techniques to find all the appearances of a runner. For improving the performance of computer vision algorithms on such research problems, a huge amount of data covering the real-world scenarios is needed.

With the increase in the number of smartphones embedded with high-quality cameras, events are easily captured and shared on the internet. Due to which there is a large availability of data. The advent of datasets of increasing scale has made a significant contribution to the advancements in computer vision. However, the scaling of such large scale datasets is hindered because of the cost and difficulty in the annotation of these large datasets with complex scenes and multiple objects. This has obstructed the progress in various deep learning tasks such as person re-identification systems that currently fail to generalize to any scenario. Another example is of text-detector systems, that fails in real-world scenarios like varying illumination, poor resolution, etc.

With the fast-paced ongoing research in machine learning, the main focus is shifting towards the applications. Deep neural networks still don't have performance saturation, and they still benefit from more data (Sun et al. 2017). Annotation of objects in the image is an equivalently time-consuming task as any other machine learning topic. According to some sources, 80\% of the AI project development time is spent on data preparation~\cite{dataStats}. ImageNet~\cite{imagenet} consists of almost 14 million sample images. It takes around 42 seconds per bounding-box by crowdsourcing using the Mechanical Turk annotation tools~\cite{crowdsourcingAF}. So, it can be imagined how much time it will take to annotate the entire dataset. Another example is of VIRAT dataset~\cite{virat}, consisting of recordings from surveillance cameras. It cost around tens of thousands of dollars to annotate the VIRAT dataset. It took around 20,000 hours to annotate object class labels and almost 5,000 additional hours for bounding box annotation in the COCO dataset~\cite{coco}.

Several other challenges persist in the collection and preparation of a large-scale representative dataset such as that of a Marathon Event, where the data is recorded using multiple cameras at different locations, containing hours of recordings of thousands of runners. Annotating such datasets with a large number of images and videos becomes difficult due to the presence of different viewpoints, varying illumination, varying camera resolution, occlusions, etc. Due to these problems, it becomes hard to annotate the runners in videos having crowded scenes and recorded at poor resolution or lighting conditions. As it is very labor-intensive to annotate such large datasets, they are generally annotated using traditional ways by crowdsourcing on platforms such as Amazon Mechanical Turk. Therefore. efficiently annotating a dataset of marathon runners is non-trivial. This area of research still needs to be explored.

Several annotation tools have been proposed in the past to accelerate the annotation process and also reducing human efforts. Almost all the tools allow the user to annotate the object of interest using a bounding box rectangle. Generally, these boxes are drawn mostly in all the video frames where the object of interest is present. Some of these tools are supported with machine learning and computer vision methods such as object detection, action recognition, object tracking, etc. to support automatic or semi-automatic annotation. However, the main drawback of such algorithms is that they are domain-specific and lacks robustness in case the complexity level in the dataset is high.

In this paper, we study the heuristics to efficiently annotate a novel in-the-wild large-scale dataset of Marathon runners consisting of 3,264 videos of almost 86 hrs of recording, covering 9,834 runners captured using 42 hand-held smartphone cameras. We explore the ways to efficiently annotate the dataset at the same time reducing the human efforts, and the cost of annotation including time as well as the budget. We investigate ways to efficiently generate bounding box annotations. We propose a novel method for aligning the identities of runners across multiple cameras. Our approach reduced the overall cost of annotation substantially by a factor of \textbf{16x} compared to the baseline method of annotation, at the same time ensuring the alignment of \textbf{93.64\%} of the total number of runners, in the cross-camera setting. 

In summary, our contributions are: We propose the annotation method of a novel dataset of Marathon runners, consisting of hours of recordings of thousands of marathon runners captured using 42 hand-held smartphone cameras. We followed a three-stage approach in the complete annotation process of the dataset. In the first stage, we study the effect of frame extraction rate in the overall time and accuracy of annotation. We study the video annotation at different fps rate to verify if it helps in reducing the overall cost.  In the second stage, we investigate different object annotation methods to annotate a runner, which is the main object of interest in our dataset. We used different computer vision-based methods such as object detection, object tracking, and box interpolation to see if they generate efficient bounding boxes in minimum time. In the final stage, we try to find ways to align the runners' identity in the cross-camera setting. We propose a runners' dashboard for the cross-camera alignment. Additionally, we show how to intelligently add noise in achieving the cross-camera alignment by using state-of-the-art person-reid methods.

% Please follow the steps outlined below when submitting your manuscript to
% the IEEE Computer Society Press.  This style guide now has several
% important modifications (for example, you are no longer warned against the
% use of sticky tape to attach your artwork to the paper), so all authors
% should read this new version.

% %-------------------------------------------------------------------------
% \subsection{Language}

% All manuscripts must be in English.

% \subsection{Dual submission}

% Please refer to the author guidelines on the CVPR 2020 web page for a
% discussion of the policy on dual submissions.

%------------------------------------------------------------------------
\section{Related work}

\textbf{Bounding box annotation.} There have been many methods proposed in the literature, for quickly generating bounding boxes. Adhikari et al. 2018, used state of the art object detector models to generate tight bounding boxes~\cite{indoorScenes}. Importantly, their object detector models are pre-trained on MS COCO~\cite{coco} dataset. Additionally, they train the model on a small subset of the dataset that is pre-labeled manually. Papadopoulos et al. 2016~\cite{weDontNeedNoBoundingBoxes} used the human-machine collaboration to generate high-quality bounding boxes. The idea is to use human verification for correcting the detections and use active learning for re-training the object detectors. One approach is to annotate only a sparse set of boxes and linearly interpolate the remaining boxes between them~\cite{labelMe}. In VATIC~\cite{vatic}, authors used shortest-path interpolation between manual annotations to generate boxes. Gygli et al. 2019~\cite{gygli2019efficient}, proposed efficiently annotating objects by clicking at the four corners of an object and speaking its label. Manen et al.~\cite{pathtrack}, proposed path supervision to annotate large-scale datasets for multi-object tracking. The authors claim that their method is efficient as it can be used to turn the watching time into annotation time. We will be investigating how we can use the object detection, box interpolation method, and object tracking techniques in efficiently generating the bounding boxes.

\textbf{Marathon Datasets.}
To the best of our knowledge, no marathon dataset has been the introduction in the literature till now. Our dataset is the first of its own kind. A similar work has been done in the past by Napolean et al. 2019~\cite{napolean2019running}, where the data is collected from a university campus marathon event. This dataset is quite a smaller version of our dataset, containing recordings of a 5km marathon race. There are a total of 262 runners captured using 9 unconstrained hand-held smartphone cameras. However, the closest resemblance of our dataset is with the person re-identification datasets~\cite{dukemtmc4reid, airport, cuhkli2013locally, cuhkli2012human, cuhkli2014deepreid, market1501}. Like person re-id datasets, our dataset also contains recordings of a number of runners appearing in multiple cameras installed at different locations. Our dataset differs from the re-id datasets in the sense that the main object of interest in our dataset is runner, whereas, in re-id datasets, all the persons appearing in the videos are the main object of interest.

% All text must be in a two-column format. The total allowable width of the
% text area is $6\frac78$ inches (17.5 cm) wide by $8\frac78$ inches (22.54
% cm) high. Columns are to be $3\frac14$ inches (8.25 cm) wide, with a
% $\frac{5}{16}$ inch (0.8 cm) space between them. The main title (on the
% first page) should begin 1.0 inch (2.54 cm) from the top edge of the
% page. The second and following pages should begin 1.0 inch (2.54 cm) from
% the top edge. On all pages, the bottom margin should be 1-1/8 inches (2.86
% cm) from the bottom edge of the page for $8.5 \times 11$-inch paper; for A4
% paper, approximately 1-5/8 inches (4.13 cm) from the bottom edge of the
% page.

% %-------------------------------------------------------------------------
% \subsection{Margins and page numbering}

% All printed material, including text, illustrations, and charts, must be kept
% within a print area 6-7/8 inches (17.5 cm) wide by 8-7/8 inches (22.54 cm)
% high.
% Page numbers should be in footer with page numbers, centered and .75
% inches from the bottom of the page and make it start at the correct page
% number rather than the 4321 in the example.  To do this fine the line (around
% line 23)
% \begin{verbatim}
% %\ifcvprfinal\pagestyle{empty}\fi
% \setcounter{page}{4321}
% \end{verbatim}
% where the number 4321 is your assigned starting page.

% Make sure the first page is numbered by commenting out the first page being
% empty on line 46
% \begin{verbatim}
% %\thispagestyle{empty}
% \end{verbatim}

%-------------------------------------------------------------------------
\section{Eindhoven Marathon Dataset}
Some datasets for marathon runners with recordings at fixed camera locations are available, however, these videos are recorded with High-Quality DSLR cameras and thus fail to generalize to real-world scenarios due to inherent domain shift. To facilitate research for in-the-wild videos, we propose a novel video dataset of marathon runners captured using handheld smartphone cameras. In total, 85 hours of videos were recorded in High Definition (HD) quality. The variability in terms of occlusion, light intensity variation, resolution, crowdedness, background clutter, and pose, etc, makes it even more challenging to learn high performing models. The dataset is also timestamped with the GPS coordinates of the location of the recording which will help keep track of the location of runners/cameras and can be useful in reconstructing the event.
   
Two marathons have been recorded, namely Half-marathon (21.5km) and Full-marathon (42km). The recording is done by the volunteers at 42 different locations covering the full-marathon track.

\subsection{Dataset Collection}
In this section, we will discuss the procedure we followed in data collection and data gathering.

\subsubsection{Organizational Logistics}
The Marathon event was organized in the city of Eindhoven, The Netherlands. Full-marathon and the half-marathon races are recorded as these events cover the entire marathon track, along with the maximum number of participants. We expected to cover approximately 10,000 marathon runners.

For coverage of the entire full marathon track, we recruited in a total of 42 volunteers. The volunteers were asked to use their personal \textit{smartphone cameras} for recording the event.
% The estimated time for recording was supposed to be an average of 2.5 hours per camera. We estimated some potential risks during the recording of the event and made a contingency plan to overcome it, as shown in Table \ref{table:riskPlan}.

% \begin{table}
% \begin{center}
% \begin{tabular}{|l|c|}
% \hline
% \textbf{Risk} & \textbf{Plan} \\
% \hline\hline
% Lack of battery backup & Power banks provided\\
% Lack of storage & Memory cards provided\\
% Travel tickets & Travel passes provided\\
% \hline
% \end{tabular}
% \end{center}
% \caption{Potential risks and their contingency plan }
% \label{table:riskPlan}
% \end{table}

\begin{figure}[ht]
                \centering
                \includegraphics[width=\linewidth, height=\linewidth]{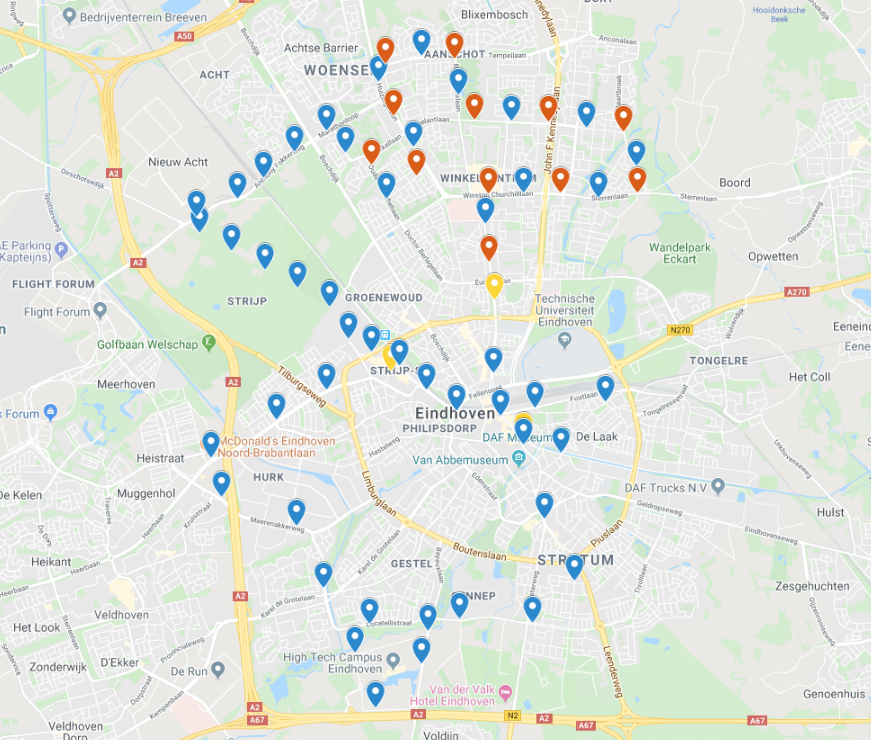}
                \caption{\textbf{A map showing the coordinates of recording locations in the marathon track.} Each Full-marathon (Blue marker) and Half-marathon (Red marker) point is at a distance of 1km from its next point of the same color. Yellow markers represents the start and the end point of the event.}
                \label{fig:map}
\end{figure}

\subsubsection{Setup}
In this section, we will discuss the recording setup and the instructions given for recording the event.

% \FloatBarrier

\begin{enumerate}
    \itemsep-0.25em
    \item Only smartphones were used for the recording and no professional cameras, DSLR's or GoPro's.
    \item Total memory space available in the devices used for recording was at least 20-30 GB
    \item Video recording resolution: 720p
    \item Frame rate: 30fps
    \item GPS enabled during recording
    \item Recording mode: LANDSCAPE
    \item Camera holders/Tripods used: None
\end{enumerate}

All the 42 volunteers were asked to record the event. A map was created and a location was assigned to each volunteer, as shown in Figure \ref{fig:map} .The recording is done at the start point and the finish point along with all the intermediate locations separated by 1 km from each other. Thirty volunteers were assigned only one coordinate, denoted by 'Point-X' and 12 volunteers were assigned two points namely 'Point-X' and 'FMP-Y'. The first 12 volunteers after recording the full-marathon at 'Point-X', shifted to their next assigned coordinate 'FMP Y' after they are done recording at the former one. The map and the assigned coordinates can be seen in Figure \ref{fig:map}. Point-X is represented by blue markers, whereas FMP-Y is represented by red-marker. The start and the end coordinates are represented by yellow markers. 

Almost all the recorded videos are timestamped and also have GPS information at the time of recording. Cameras were hand-held during the recording, and no stands/tripods are used. Therefore, the recordings are shaky rather than static.

% \begin{figure}[hbt!]
%                 \centering
%                 \includegraphics[width=\linewidth, height=\linewidth]{images/Map.png}
%                 \caption{\textit{Coordinates assignment: Each Full-marathon(Blue marker) and Half-marathon(Red marker) point is at a distance of 1km from its next point of the same color. Yellow markers represents the start and the end point of the event.}}
%                 \label{fig:map}
% \end{figure}
% % \FloatBarrier

\subsubsection{Execution}
\begin{figure}[hbt!]
                \centering
                \includegraphics[width=\linewidth, height=0.5\linewidth]{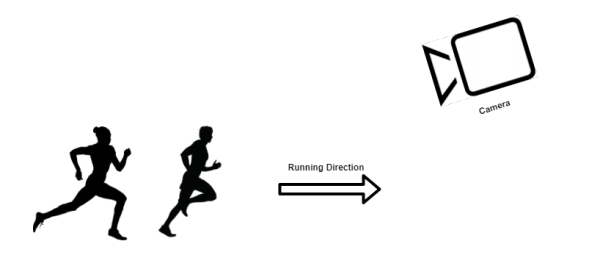}
                \caption{\textbf{Illustration of how video recording was configured.} The recording is done from the front side to capture the bibId attached to the uniform of the runner. The recording is started when a runner is visible for the first time and is done until the runner leaves the camera frame~\cite{napolean2019running}.}
                \label{fig:recDirection}
\end{figure}

The recording was started whenever a runner was visible for the first time. The recording was stopped when no runner was visible in the frame. The recording was done from the front side of the runner so that bibId was visible and could be recorded in the video. The backside of the runner was not recorded as there was no bib-number on the back of the runner. Our setup in terms of camera position and the athletes resembles Figure \ref{fig:recDirection}, in which the runners ran towards the camera. Initially, there was a considerable distance between the camera and the runners, but the runners progressively got closer to the camera.

% \FloatBarrier

\subsection{Data Gathering}
\subsubsection{Collecting the recorded data}
After the videos have been recorded, they needed to be collected at one single storage point. For that, we used a webserver with a storage capacity of 4TB. A website was created for the volunteers to upload their collected data onto the webserver. The participating athletes were also asked to upload the strava details, that is activity and performance tracking details of athletes, that they have collected during their run.

\subsubsection{Web Scraping}
\begin{figure}[ht]
    \centering
    % \subfloat[Sample images scraped from official website]
    {\includegraphics[width=0.4\linewidth,height=0.4\linewidth]{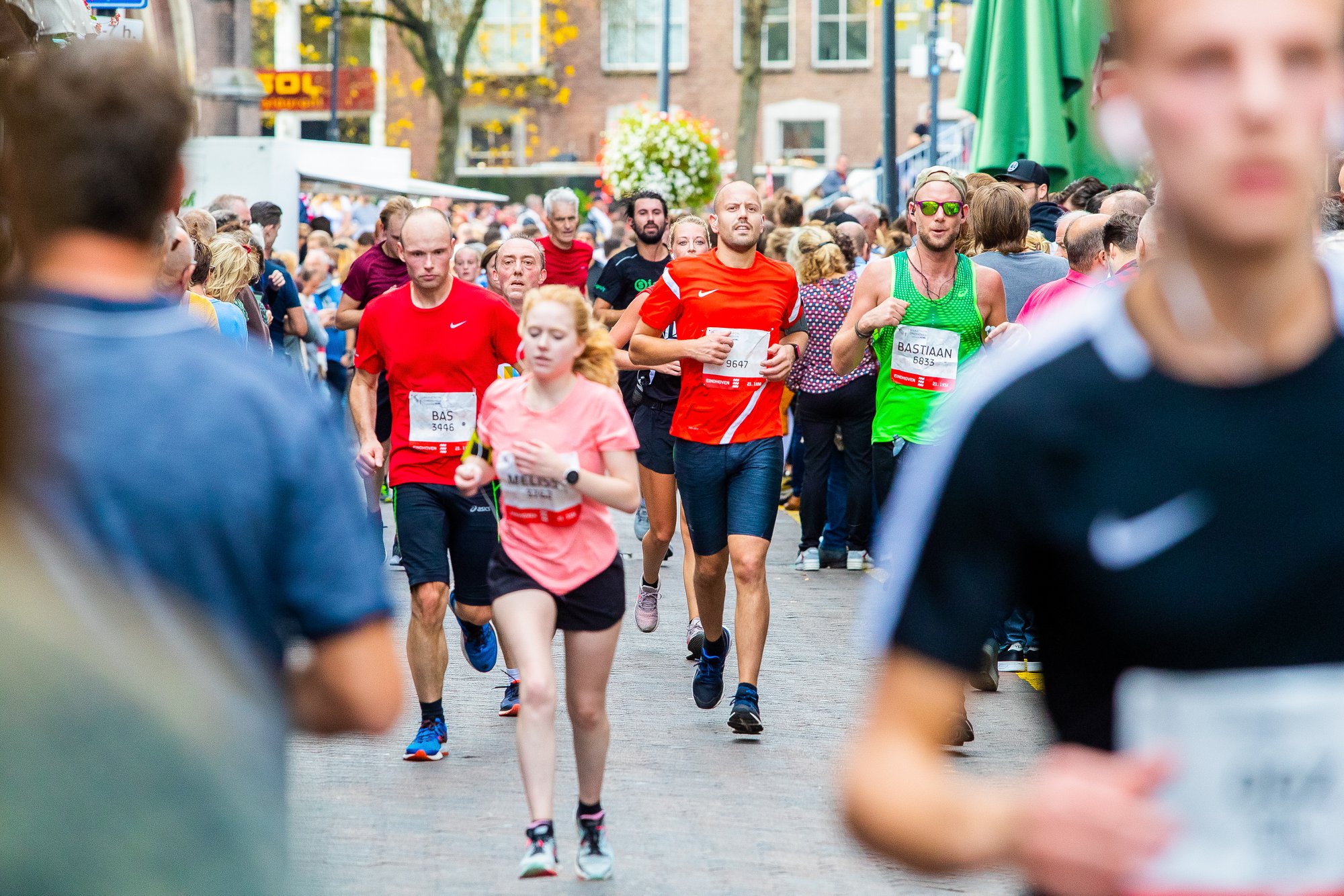} }
    {\includegraphics[width=0.4\linewidth,height=0.4\linewidth]{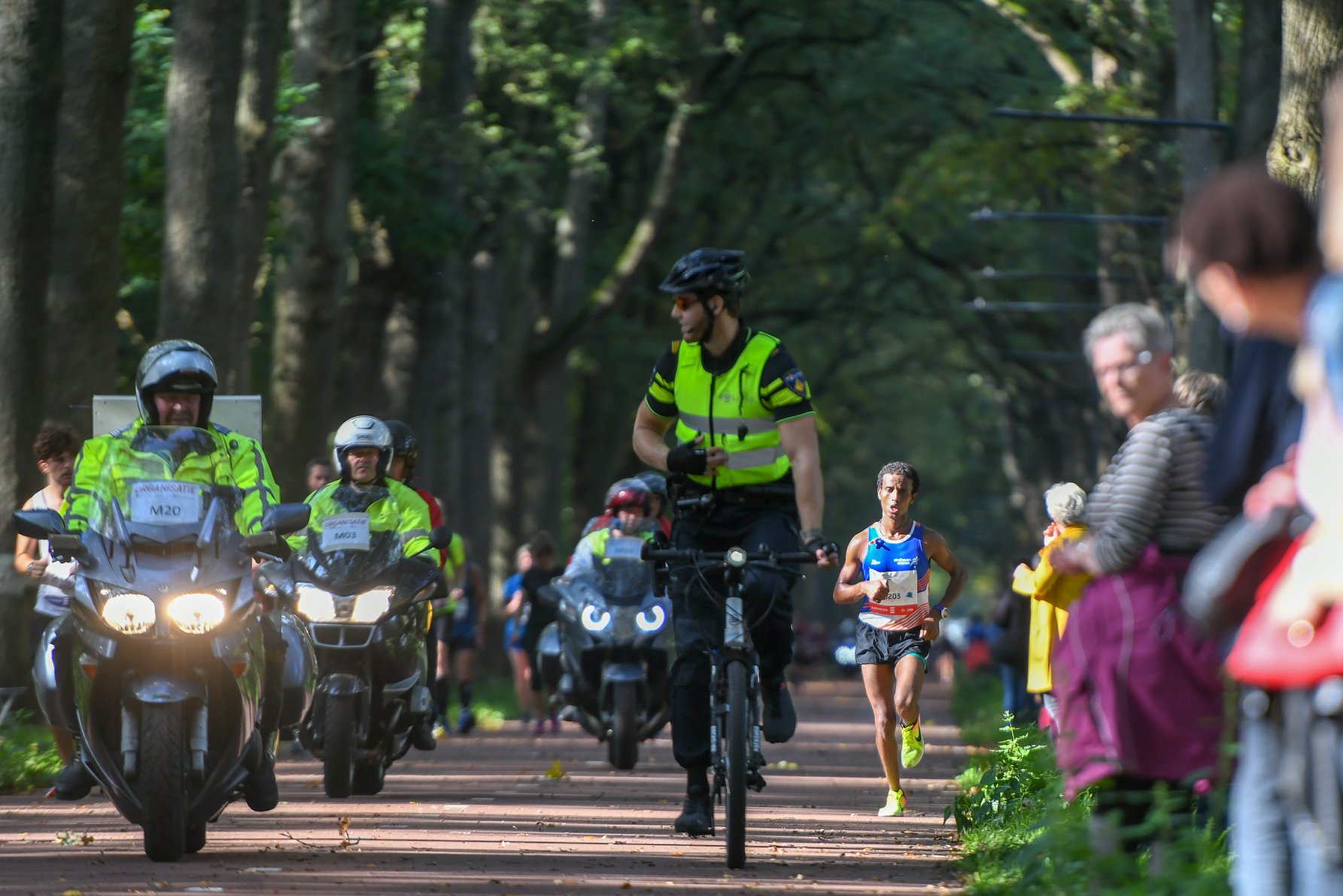} }
    \caption{\textbf{Sample images of marathon-event scraped from official website~\cite{officialWebsite}.} These images are not part of the dataset, and are publicly available on the website. }%
    \label{fig:sampleScrapedImages}%
\end{figure}

To collect the detailed information of all the participants of the full-marathon and half-marathon, web-scraping was done from the official website of the Marathon Eindhoven event~\cite{officialWebsite}. Information such as runner name, runner bib-id, age, country, finish times after distances(in meters) 5k, 10k, 15k, 20k, 25k, 30k, 35k, 40k, and 42k are collected. A complete list of field names is mentioned in Table \ref{table:fieldNames}. Event images were also collected from the official website~\cite{officialWebsite}, sample images can be seen in Figure \ref{fig:sampleScrapedImages}. More information on data scraping is provided in \ref{apdx:webscraping}.

\subsubsection{Collecting Video Metadata}
After recording the data and collecting it on a common platform, the metadata of all the videos is read using a python library \textit{exif Tool}~\cite{exifTool}. Using exifTool~\cite{exifTool} we collected information of a video such as \textit{FileName, FileSize, FileType, Duration VideoFrameRate, ImageSize, TrackCreateDate, GPSCoordinates}. A sample of metadata of a collected video is shown in Table \ref{table:metadataTable}.

% \begin{table}
% \begin{center}
% \begin{tabular}{|l|c|}
% \hline
%          FileName & VID\_20191013\_114355.mp4\\\hline
%          FileSize(MB) & 21.92\\\hline
%          FileType & MP4\\\hline
%          Duration & 14.34\\\hline
%          VideoFrameRate & 30\\\hline
%          ImageSize & 1280x720\\\hline
%          TrackCreateDate & 2019:10:13 09:43:55\\\hline
%          GPSCoordinates & 51.4839 5.4642\\
% \hline
% \end{tabular}
% \end{center}
% \caption{\textbf{A sample of collected video metadata.} Video metadata such as FileName, FileSize, FileType, Duration VideoFrameRate, ImageSize, TrackCreateDate, GPSCoordinates are collected.}
% \label{table:metadataTable}
% \end{table}

\begin{table}
\begin{center}
\begin{tabular}{l c}
\toprule
\textbf{Video Metadata} & \textbf{Value}\\
\midrule
         FileName & VID\_20191013\_114355.mp4\\
         FileSize(MB) & 21.92\\
         FileType & MP4\\
         Duration & 14.34\\
         VideoFrameRate & 30\\
         ImageSize & 1280x720\\
         TrackCreateDate & 2019:10:13 09:43:55\\
         GPSCoordinates & 51.4839 5.4642\\
\bottomrule
\end{tabular}
\end{center}
\caption{\textbf{A sample of collected video metadata.} Video metadata such as FileName, FileSize, FileType, Duration VideoFrameRate, ImageSize, TrackCreateDate, GPSCoordinates are collected.}
\label{table:metadataTable}
\end{table}

\subsection{Data Analysis}
\begin{table}[ht]
\begin{center}
\begin{tabular}{ c  c }
\toprule
\multicolumn{2}{c}{\textbf{Statistics}} \\
\midrule
         Total no. of cameras & 42\\
         No. of videos recorded & 3,264\\
         Mean duration of videos  & 94.13 sec\\
         Standard Deviation & 246.78\\
         Total duration of recording & 85.34 hrs\\
         Total frames in dataset & 9,216,813\\
         Runners in full-marathon & 2,423\\
         Runners in half-marathon & 7,411\\
         Total runners & 9,834\\
\bottomrule
\end{tabular}
\end{center}
\caption{\textbf{Statistics of the Eindhoven Marathon dataset.} Total 9,834 runners are captured using 42 cameras. The dataset consists of 3,264 videos of 85.34 hrs of recording. The mean duration of the videos is around 95 seconds.}
\label{table:metadataTable2}
\end{table}

\begin{figure}[ht]
                \centering
                \includegraphics[width=\linewidth, height=0.5\linewidth]{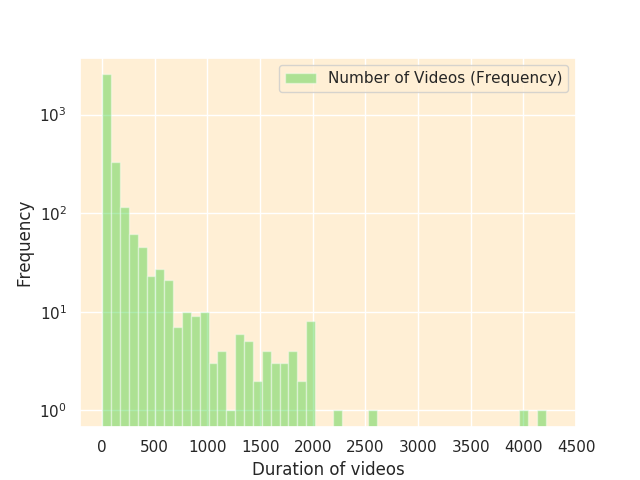}
                \caption{\textbf{Analyzing video duration vs the total number of videos of that duration (frequency).} It can be seen from the figure that the mean duration of the videos is centered around 95 seconds}
                \label{fig:metadataAnalysis}
\end{figure}

The collected metadata is analyzed and mentioned in Table \ref{table:metadataTable2} and Figure \ref{fig:metadataAnalysis}. As can be seen in Figure \ref{fig:metadataAnalysis}, most of the videos are short in length with average duration around 95 seconds. All videos are recorded at 30fps and at High-Definition (HD) resolution. Therefore, there are a total of 9,216,813 frames in the dataset. The total video duration is of 85.34 hrs. A total of 9,834 runners are covered including full-marathon and half-marathon.

\subsection{General Data Protection Regulation (GDPR)}
The data is recorded and collected, in adherence to the GDPR guidelines~\cite{GDPR}. In general, the main focus of data recording was the marathon event which is organized publicly every year. None of the data subjects is focused or recorded individually. Although, the participants gave their consent of data recording and sharing to the MyLaps~\cite{mylaps}, an official organizer of the Eindhoven Marathon event, that is also working in collaboration with the Computer Vision Lab~\cite{cvlab} at the university. We adhered to the privacy regulations, by blurring/hiding the faces of the people appearing in the images. Also, the data is not uploaded or shared publicly.

\section{Methodology}

\begin{figure*}[ht]
                \centering
              \includegraphics[width=\linewidth]{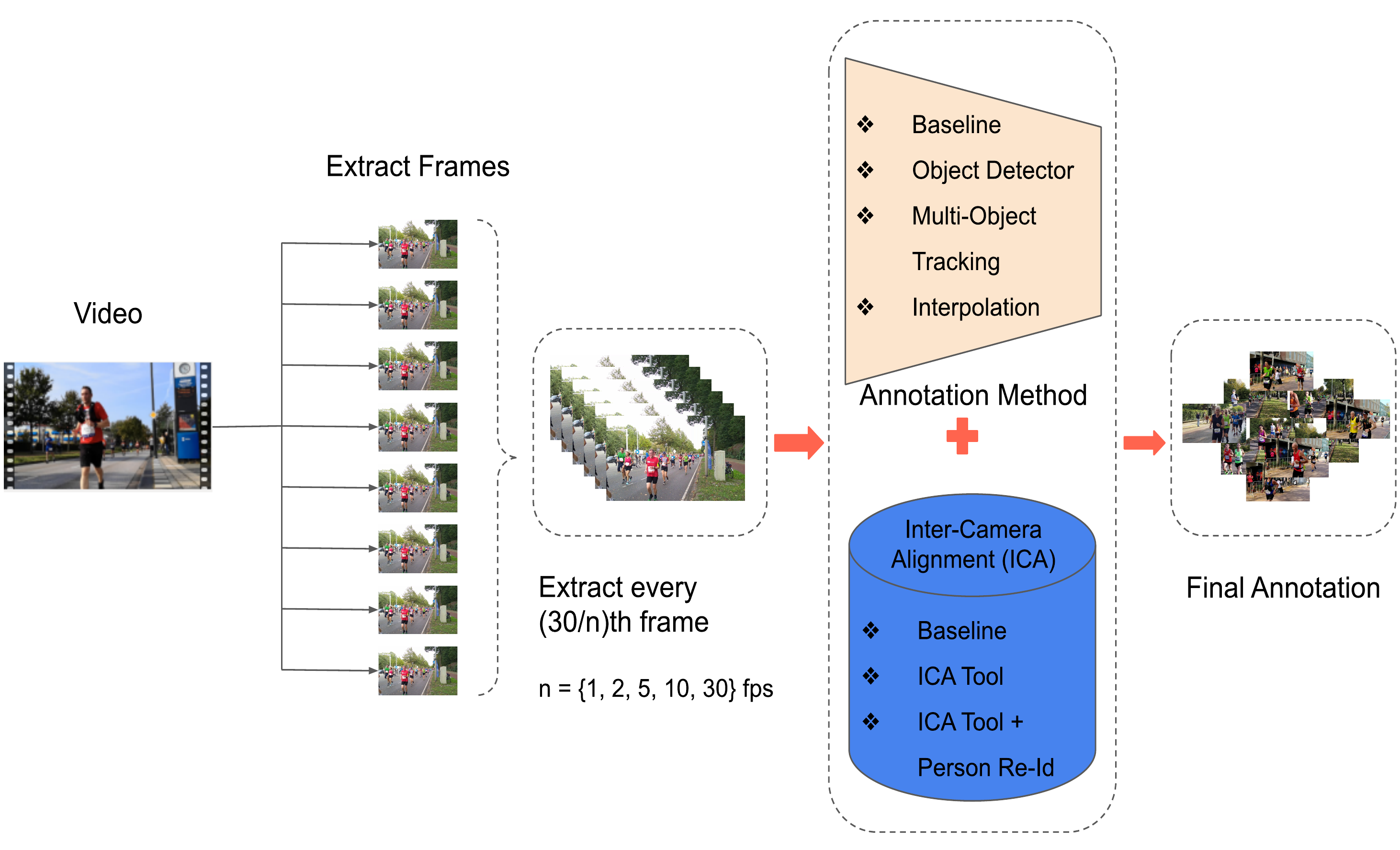}
                \caption{\textbf{Overall annotation pipeline.} The videos are recorded at the rate of 30 frames per second (fps). Firstly, all the frames are extracted from videos. Then, we analyze the effect of \textit{fps} on annotation time, by sampling frames at different fps. The sampled data is then annotated using bounding box regression, and the runners are aligned across multiple disjoint cameras, to get final annotation.}
                \label{fig:pipelineDiagram}
\end{figure*}

In this section, we will discuss the general annotation procedure and the pipeline that is used in our experiments. Then we will cover the method used to derive a sample dataset, followed by a discussion over the data collection procedure and the Inter-Camera Alignment (ICA) tool. Lastly, we will elaborate on the metrics that are used in evaluating the performance of the methods.

\subsection{Annotation Procedure}

\textbf{Video Annotation.} Annotating videos is similar to image annotation. It involves mainly two steps: i) Extracting the frames from videos, ii) Annotation of individual frames.

The overall annotation pipeline is shown in Figure \ref{fig:pipelineDiagram}. Firstly, frames are extracted at a specific extraction rate. Then, each runner in individual frame are annotated using the proposed annotation method. Later, the identities of the runners are aligned using the proposed method for inter-camera alignment, to get the final annotations.

\textbf{Manually annotating runners.} The main object of interest here is the runner. We used the basic bounding box annotation around the runners. Every runner is assigned a unique runner id (\textit{bibId}) which is attached on the runner's chest. The bibId of the runner is used as the label which is manually added by the annotator. Firstly, a runner is selected, then we try to find a frame in which the runner's bibId is visible. We then move back to the frame number where the runner is visible for the first time in the camera. We start putting bounding boxes around the runner in all the frames in which he/she is visible until the runner leaves the camera frame. We followed runner-wise annotation instead of frame-wise annotating all objects of interest.

\subsection{Sample Dataset Creation.}
\label{sec:sampleDataset}
The large size of the unannotated dataset makes it difficult for validating the experiments. Hence, a sample dataset is required, that is a true representative of the actual dataset. A smaller version of the dataset is created by down-sampling the original one and covering different scenarios present in different video locations. 

\begin{figure*}
\begin{center}
\subfloat[Resolution]{{\includegraphics[width=0.19\textwidth]{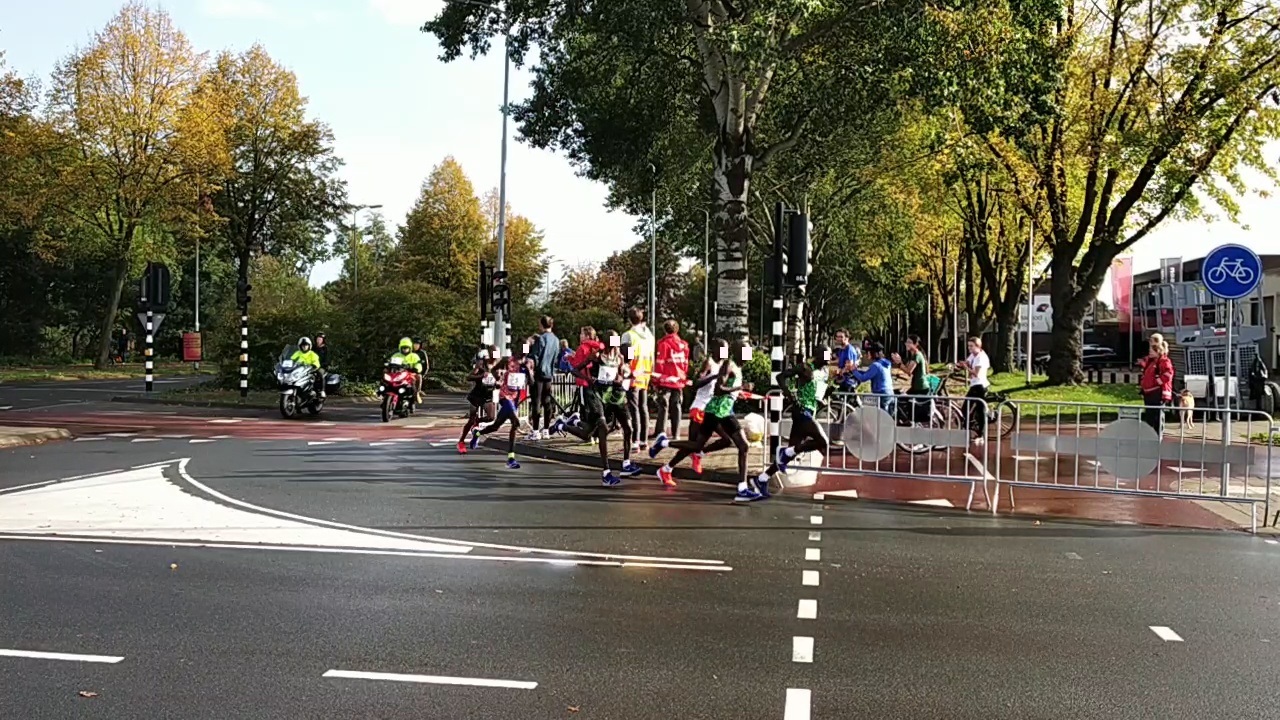} }}%
    \hfill
    \subfloat[Lighting Condition]{{\includegraphics[width=0.19\textwidth]{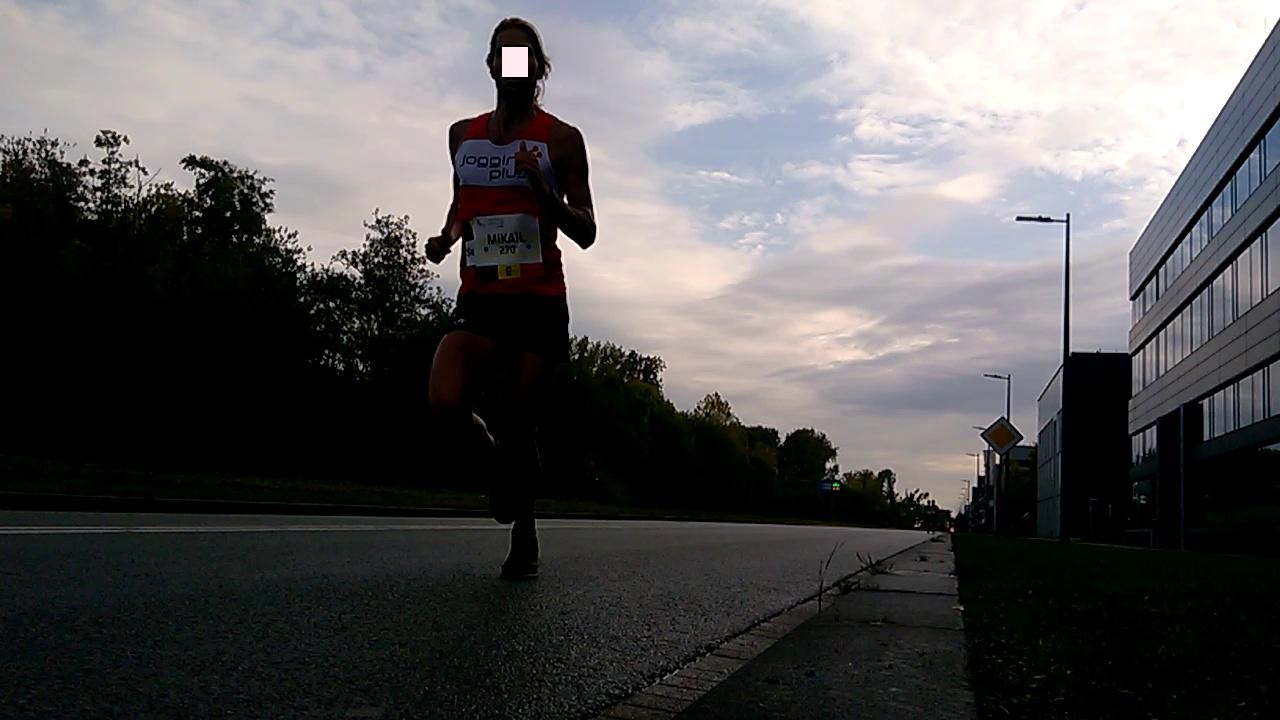} }}%
    \hfill
    \subfloat[Recording Angle]{{\includegraphics[width=0.19\textwidth]{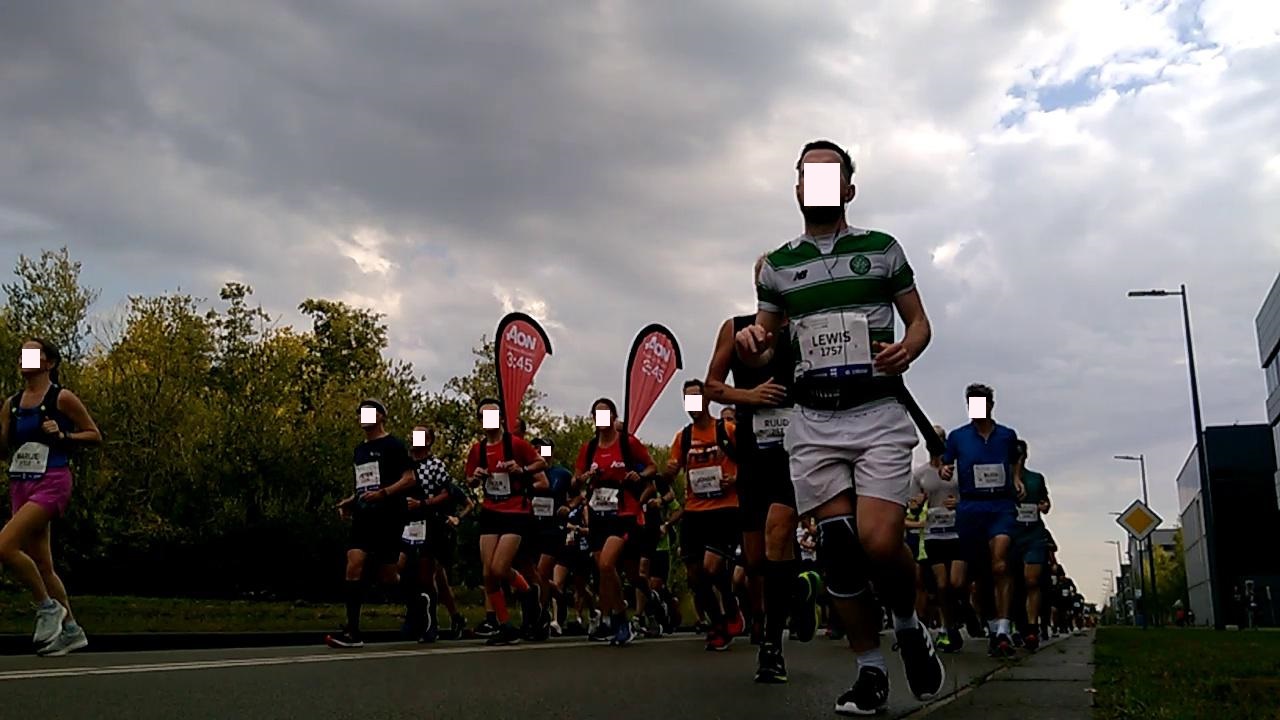} }}%
    \hfill
    \subfloat[Occlusion]{{\includegraphics[width=0.19\textwidth]{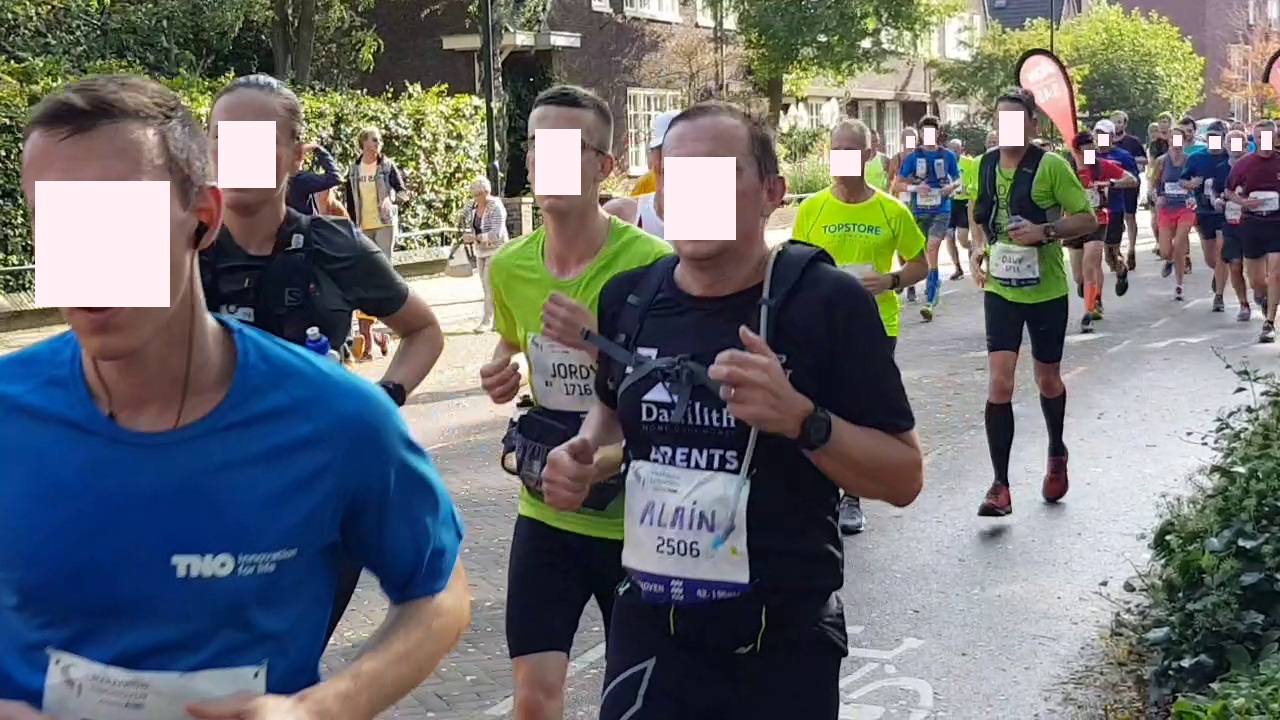} }}%
    \hfill
    \subfloat[Crowdedness]{{\includegraphics[width=0.19\textwidth]{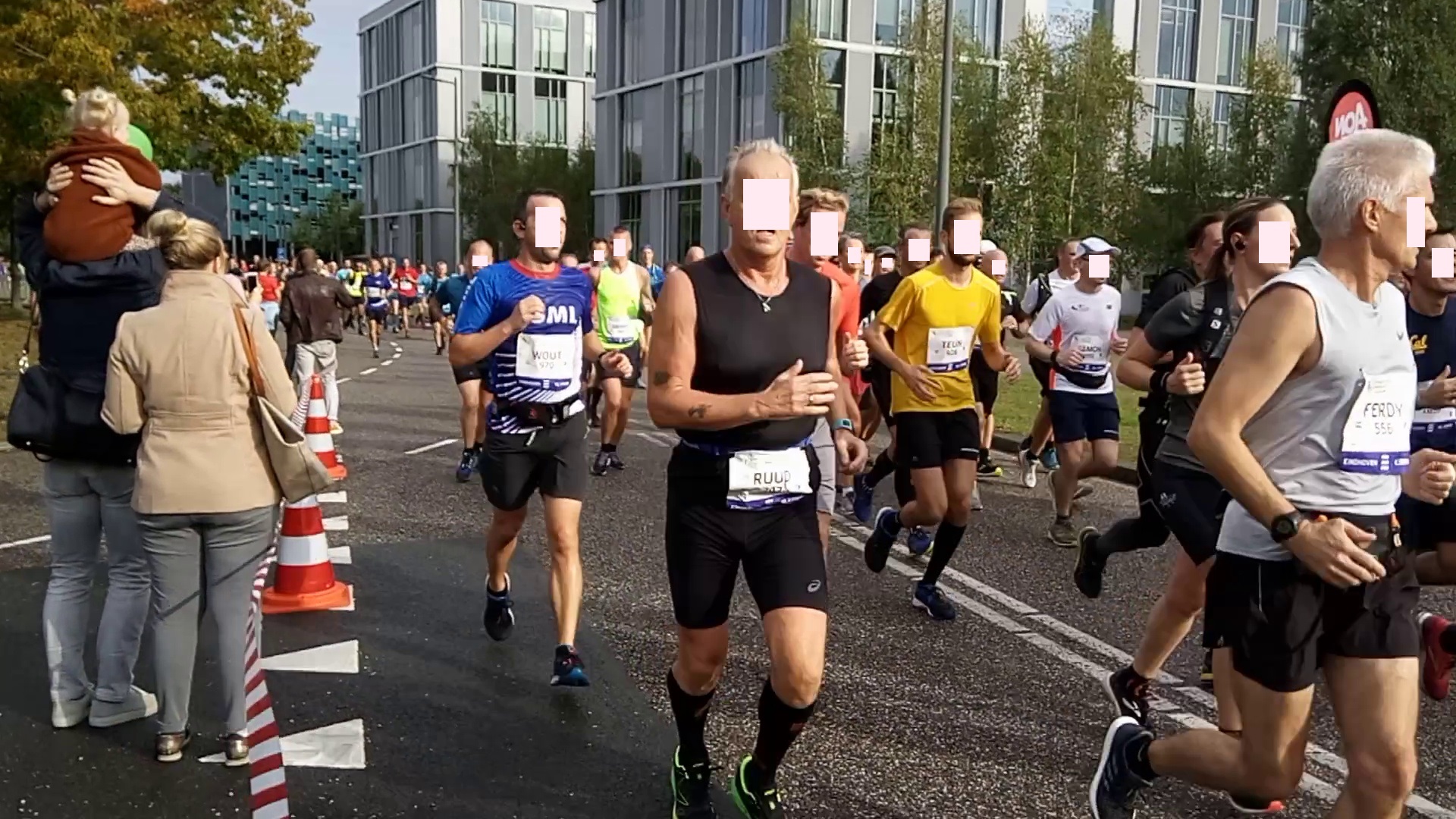} }}%
\end{center}
   \caption{\textbf{The five main scenarios present in the dataset} are i) resolution, ii) lighting condition, iii) recording angle, iv) occlusion, v) number of crowded videos.}
\label{fig:scenarios}
\end{figure*}

\textbf{Scores Assignment for Locations.}\label{subsec:scenarios} Firstly, variations in different scenarios are analyzed at all the recording locations. The scenarios are divided into the following five categories: \textit{i) the resolution, ii) lighting condition, iii) recording angle, iv) occlusion, v) number of crowded videos}. Some examples of these scenarios are shown in Figure \ref{fig:scenarios}.

\begin{table*}[ht]
\begin{center}
\begin{tabular}{ p{1cm}  p{2cm}  p{2cm}  p{5cm}  p{2cm}  p{3cm} }
\toprule
         \textbf{Scores} & \textbf{Lighting} & \textbf{Resolution} & \textbf{Recording Angle} & \textbf{Occlusion} & \textbf{\# Crowded Videos}\\\midrule
         1 & Very poor & Very poor & Front+Side+Static+Upside-Down & Very high & Very large\\
         2 & Poor & Poor & Side & High & Large\\
         3 & Moderate & Moderate & Front+Downside+Static & Moderate & Moderate\\
         4 & Good & Good & Front+Side & Low & Less\\
         5 & Very good & Very good & Front & Very low & Very less\\
\bottomrule
\end{tabular}
\end{center}
\caption{\textbf{Different scenarios in the dataset and their score distribution.} All video locations are analyzed and penalized as per the scenarios of the video recording at that location. A location score is the sum of scores in different scenarios.}
\label{table:scoreDistribution}
\end{table*}

Different scenarios and their score distribution is shown in Table \ref{table:scoreDistribution}. The total score of a location is the sum of scores in individual categories of different scenarios. Location score $S_{location}$ is given by,

\begin{equation}
\begin{split}
    S_{\text{location}} = S_{\text{lighting}} + S_{\text{resolution}} + S_{\text{recordingAngle}} \\ 
                  + S_ {\text{occlusion}} + S_{\text{noOfCrowdedVideos}}.
\label{eq:totalScore}
\end{split}
\end{equation}

Consider an example location 'X' having videos recorded in poor lighting conditions(2), with good resolution(5), recording angle is front(5), moderate occlusion(3) \& number of videos having too many crowded runners is moderate(3), then the total score is 
$S_{X}= 2 + 5 + 5 + 3 + 3=18$.

\begin{figure}[ht]
                \centering
                \includegraphics[width=\linewidth]{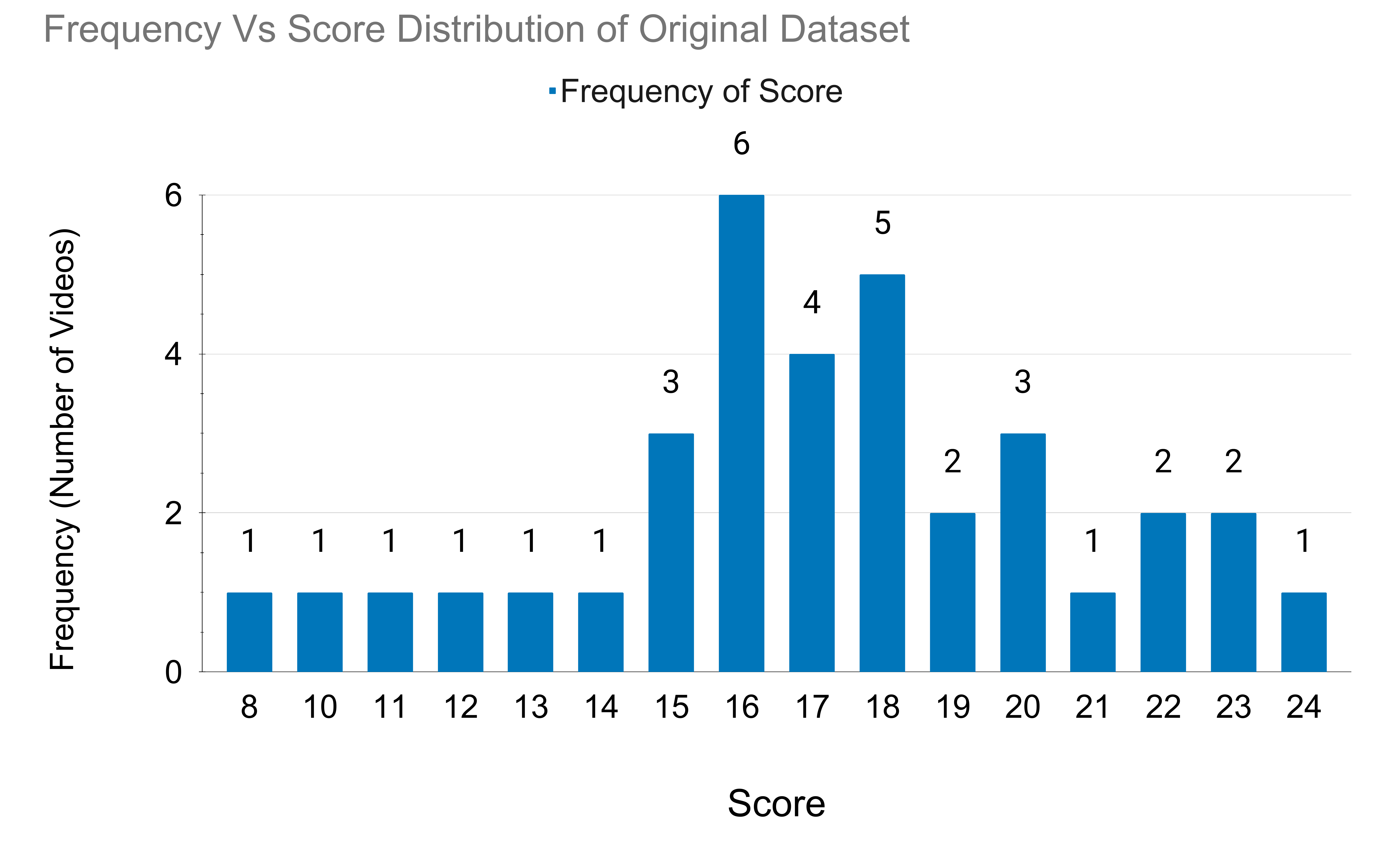}
                \caption{\textbf{Score distribution of the entire dataset.} Frequency represents the number of locations with a given score. The distribution of the dataset is a Gaussian.}
                \label{fig:freqScoreDistribution}
\end{figure}

All the 42 recorded locations are given overall score S = [0, 25], where for each of the above-mentioned categories we can have a score in the range [0, 5]. For more details, refer to \ref{apdx:ksTest}. The frequency vs score distribution of the entire dataset is shown in Figure \ref{fig:freqScoreDistribution}. There are a total of 16 different score values ranging from 8 to 24. Higher scores imply better recording scenarios in videos.

\textbf{Kolmogorov-Smirnov (KS) Test.} The Kolmogorov–Smirnov test is used to test the similarity of two underlying one-dimensional probability distributions.

The two-sample K–S test is one of the most useful and general non-parametric methods for comparing two samples, as it is sensitive to differences in both location and shape of the empirical cumulative distribution functions of the two samples.

For larger sample sizes, the approximate critical value $D_{\alpha}$ is given by the Equation \ref{eq:d_alpha},

\begin{equation}
    D_{\alpha } = c{(\alpha)}\sqrt{\frac{n_{1} + n_{2}}{n_{1}n_{2}}},
\label{eq:d_alpha}
\end{equation}

where, $n_{1}$ and $n_{2}$ are the sample sizes of the two distributions and $\alpha$ and $c{(\alpha)}$ are the coefficients given by Table~\cite{kstest}. More information can be found in Appendix \ref{apdx:ksTest}.

\textbf{How to select locations?} To select the desired number of locations for the sample dataset, we use Kolmogorov-Smirnov Test for distribution similarity matching. The score distribution of both the sample dataset and the original dataset is assumed to be Gaussian. Here, we randomly select 6 scores out of a total of 16 and matches the similarity of the selected scores with the original dataset scores. The K-S test calculates the distance between the two distributions, and the sample distribution of the randomly selected 6 score values, having the minimum distance from the actual data distribution is selected.

\begin{figure}[ht]
                \centering
                \includegraphics[width=\linewidth]{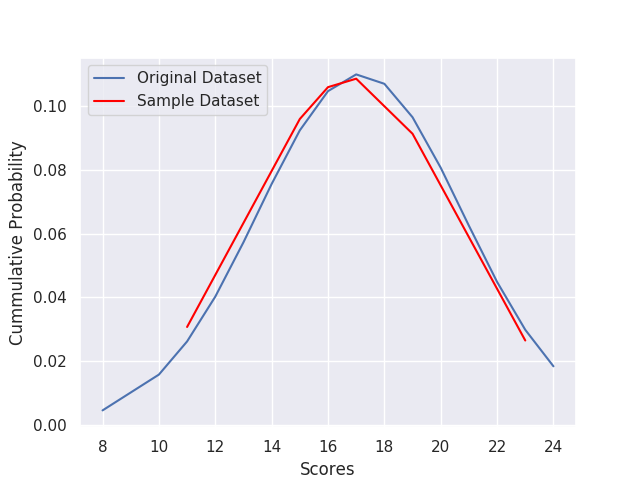}
                \caption{\textbf{The graph shows the probability distribution of the scores of the original and the sampled dataset.} The sample score values [11, 15, 16, 17, 19, 23] is having distribution that fits nicely with the actual dataset score distribution.}
                \label{fig:distributionSimilarity}
\end{figure}

We selected score values \{11, 15, 16, 17, 19, 23\} for the sample dataset. The distribution of the selected values and the original dataset is shown in Figure \ref{fig:distributionSimilarity}. Now, 6 locations having these scores are selected. In all, 60 videos are collected, corresponding to the top 20 and middle 3 runners. It is made sure that all these runners appear in all the 6 sampled video locations as it's necessary for the validation of the cross-camera alignment. However, due to unavoidable circumstances, there can be runners other than the selected ones, appearing in each sampled location.

\subsection{Data Collection}
\label{dataCollection}
For achieving the cross-camera alignment of runners, our methods require the information of every runner that finished the race. For this, we scraped the images and information of every runner from the official website of marathon~\cite{officialWebsite}.

\textbf{Scraping runners' information.} The data is scraped from the official website~\cite{officialWebsite} of the event using Beautiful soup~\cite{beautifulsoup} and Selenium Web-driver~\cite{selenium}. For every participant, information such as name, bibId, gender, country, finish times, etc is retrieved. More information is mentioned in appendix \ref{apdx:webscraping}. Apart from this, we also have information corresponding to the time taken by every runner to reach every 5 km distance intervals. 

\begin{figure}[ht]
    \centering
    % \subfloat[Sample images scraped from official website]
    {\includegraphics[width=0.3\linewidth,height=0.4\linewidth]{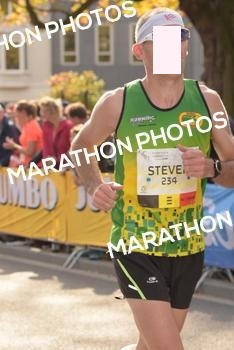} }
    {\includegraphics[width=0.3\linewidth,height=0.4\linewidth]{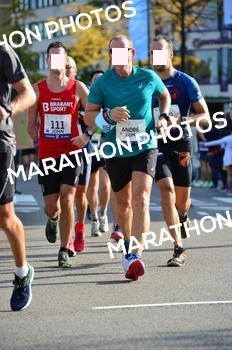} }
    {\includegraphics[width=0.3\linewidth,height=0.4\linewidth]{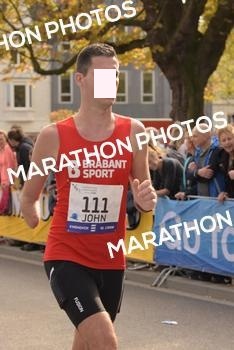} }
    \caption{\textbf{Sample images of marathon-event scraped from official website~\cite{officialWebsite}.} All the scraped images contains a watermark text.}%
    \label{fig:scrapedImages}%
    
\end{figure}

\textbf{Scraping runners' images.} We also scraped images of all the runners available on the website~\cite{officialWebsite}. The image resolution is $233 \times 350$ pixels. On average, 40 images per runner are scraped. Sample images can be seen in Figure \ref{fig:scrapedImages}.

\subsection{Inter-Camera Alignment (ICA) Tool}
\label{subsec:icaTool}
This section outlines the computation of runners' timeline and the creation of an interactive Inter-Camera Alignment tool for alignment of runners in cross-camera setting.

\textbf{Computing the runner's timeline.} For ICA, it is important to have the athlete's trajectory, to know the whereabouts of the runner at different locations. As discussed in Section \ref{dataCollection}, we have scraped the data corresponding to every runner who finished the race, which also includes the time taken by a runner to reach distances of 5 km interval. The event is recorded at 42 different locations, each separated by 1 km distance, covering the full-marathon track. So we need time values of every runner to reach at these 42 locations. These time values are calculated using \textit{variable average speed concept}.

\textbf{Variable average speed.} In general, the average speed of a runner is given by Equation \ref{eq:Vavg},

\begin{equation}
    \text{V}_{\text{avg}} = \frac{\text{Total distance travelled}}{\text{Total time taken}} = \frac{42}{\text{t}_{\text{finish}}}.
\label{eq:Vavg}
\end{equation}

\begin{figure}[ht]
                \centering
                \includegraphics[width=\linewidth]{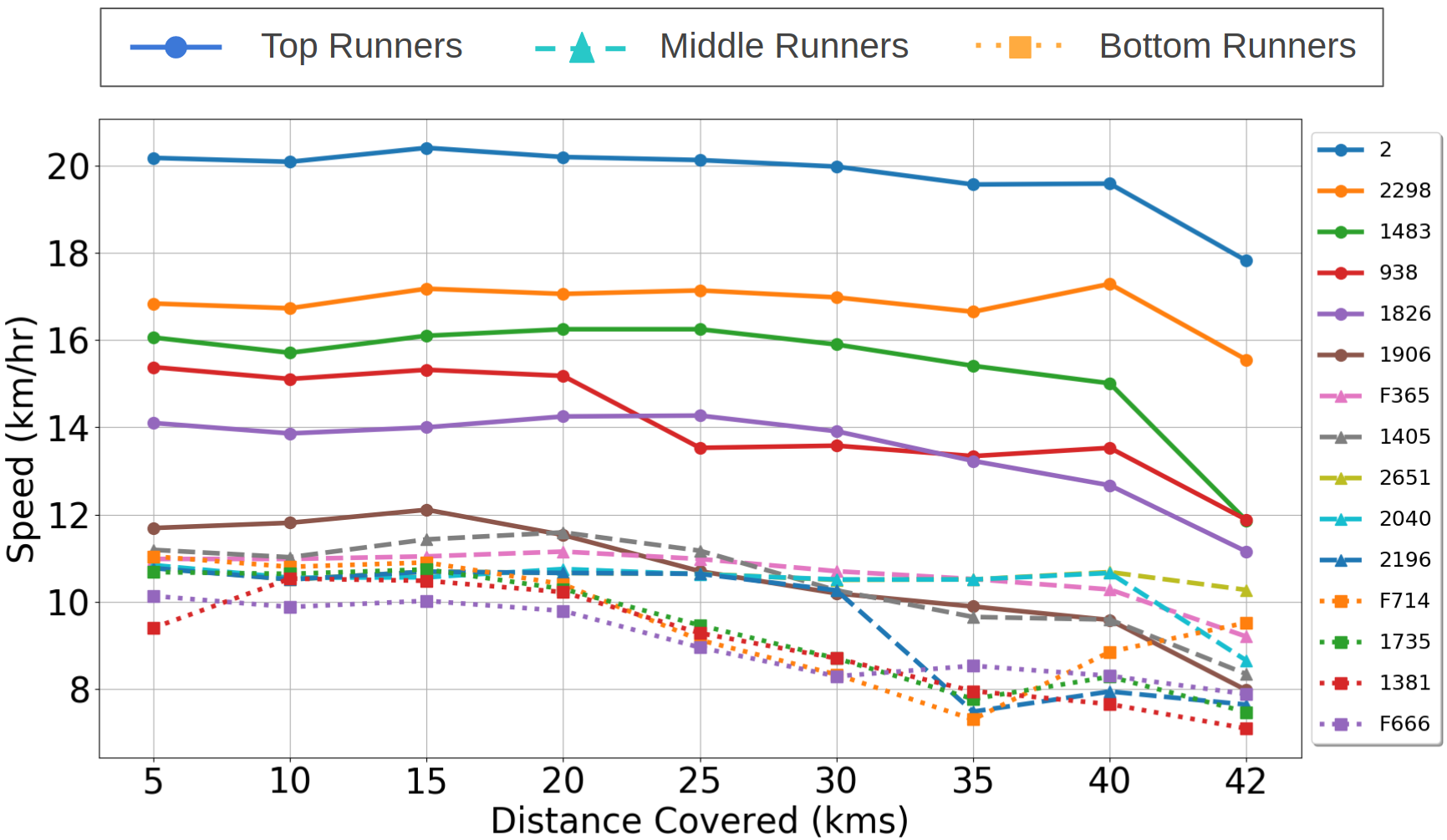}
                \caption{\textbf{Speed variations of top 5, middle 5, and bottom 5 runners, chosen randomly from top 800, middle 800, and bottom 800 full-marathon runners respectively.} It can be inferred that runners' speed is not constant throughout the run. Also, the top runners run at a more constant pace than middle and bottom runners.}
                \label{fig:runnerSpeedAnalysis}
\end{figure}

But the average speed concept is valid in case the runners is always running with constant speeds in different segment. But as can be seen in Figure \ref{fig:runnerSpeedAnalysis}, only top runners are running with more continuous pace than middle and bottom runners. Therefore, we use variable average speed concept.

It means, using different average speeds for a different segment of the race. As shown in Figure \ref{fig:variableAvgSpeed}, the time taken (in green) by the runner to reach every 5km distance is retrieved from the official website~\cite{officialWebsite}. There are cases where the time value is missing for some intermediate checkpoints (in red). In this case, the variable average speed is calculated as follows:

\begin{figure}[ht]
                \centering
                \includegraphics[width=\linewidth]{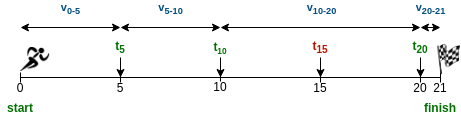}
                \caption{\textbf{Using the variable average speed concept to compute the complete timeline of a runner.} Time values in green color are retrieved from the official website, whereas the value in red is missing. }
                \label{fig:variableAvgSpeed}
\end{figure}

Consider a segment of a track i $\rightarrow $ j, where i,j are checkpoints and i $<$ j. A segment is defined as the two consecutive checkpoints for which the time values are available. Let's calculate the time ($t_{x}$) taken by a runner to reach a point $x$, where $i < x < j$. For a given segment the variable average speed is given by Equation \ref{eq:varAvgSpeed},

\begin{equation}
    \text{V}_{\text{i-j}} = \frac{\text{d}_{\text{i-j}}}{\text{t}_{\text{i-j}}} = \frac{\text{d}_{\text{0-j}} - \text{d}_{\text{0-i}}}{\text{t}_{\text{0-j}} - \text{t}_{\text{0-i}}}, \hspace{10mm} \text{where } i < j.
\label{eq:varAvgSpeed}
\end{equation}

Now, $\text{V}_{\text{i-j}}$ is used to compute the time ($t_{x}$) taken by runner to reach point $x$, where $i < x < j$, 

\begin{equation}
    \text{t}_{\text{x}} =  \frac{\text{d}_{\text{i-x}}}{\text{V}_{\text{i-j}}} = \frac{\text{d}_{\text{0-x}} - \text{d}_{\text{0-i}}}{\text{V}_{\text{i-j}}},
\label{eq:timeCalculation}
\end{equation}

Then the time ($t_{x}$) taken by a runner to reach the point $x$ is given by Equation \ref{eq:timeCalculation}. This way, for every runner we calculate the time taken by him/her to reach all the 42 locations.

\textbf{Runners' dashboard.} For aligning the identities of runners across different camera locations, a dashboard of runners is created. Runners' information such as name, bibId, and timeline of the entire event are some of the main features of the tool. It also incorporates a complete timeline of every runner computed using the variable average speed concept. The interactive tool provides features for searching and sorting over all the fields, view timeline diagrams of runners, thereby helping in quickly finding a runner's location at a particular point in time. There are separate tabs for full-marathon and half-marathon runners. There's also a tab containing rules of inter-annotator agreement.

\begin{figure*}[ht]
    \centering
    % \subfloat[Sample images scraped from official website]
    {\includegraphics[width=0.48\linewidth,height=0.3\linewidth]{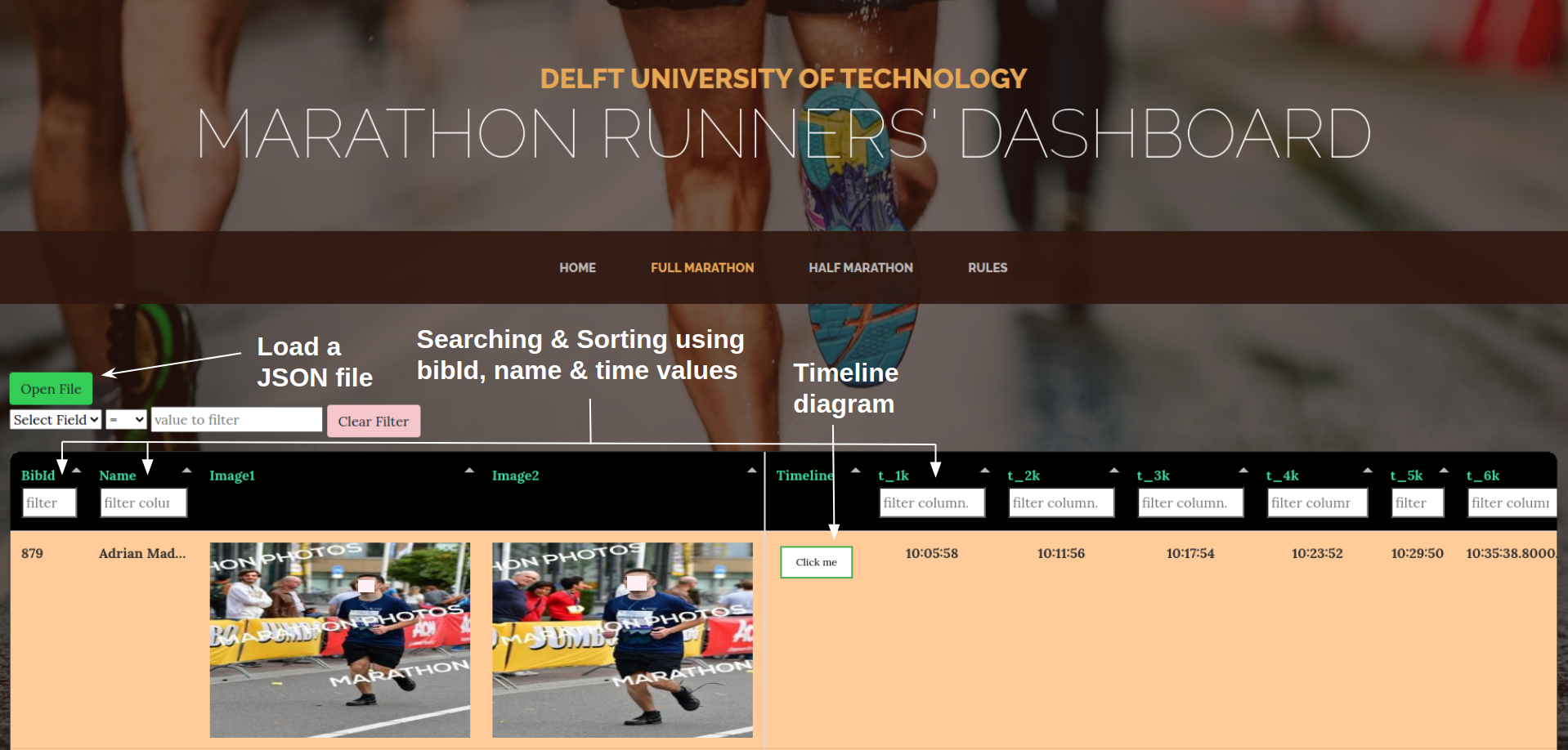} }
    {\includegraphics[width=0.48\linewidth,height=0.3\linewidth]{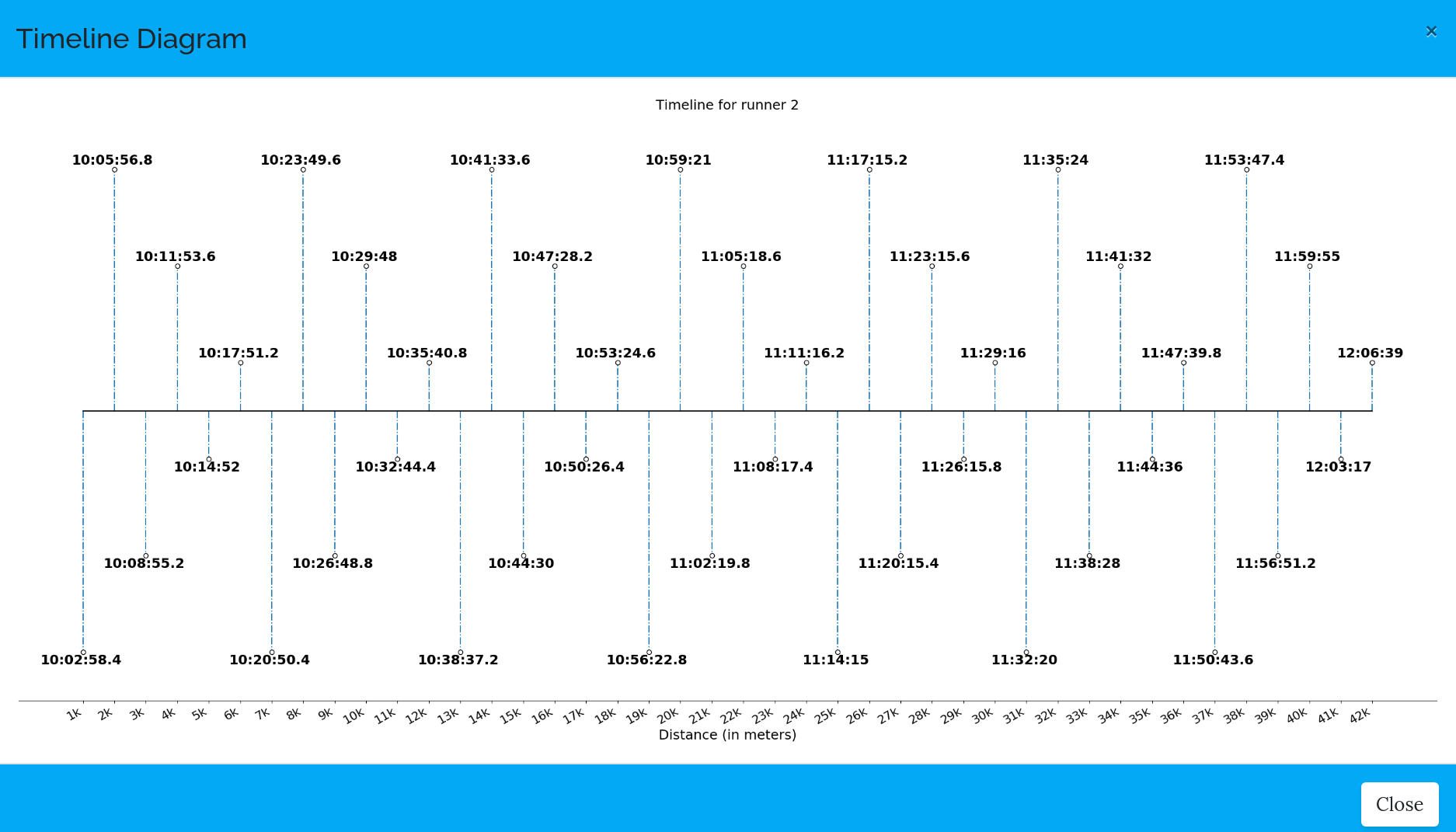} }
    \caption{\textbf{An outlook of the Inter-Camera Alignment tool developed to align the runners across different camera locations.} The tool incorporates features such as runners' names, bibId, time values, timeline diagram, etc. The timeline diagram helps the annotator in quickly checking the time taken by the runner to reach different locations.}%
    \label{fig:ICATool}%
    
\end{figure*}

The images of runners included in the tool are scraped from the official website~\cite{officialWebsite} of the event. Only two images per runner are included, to help the annotators in visually recognizing the runner. A runner can also be searched using the name or part of a name, using its bibId or part of a bibId. This partial name or bibId searching facility helps in reducing the search space. The partial searching feature is also valid for time values. All the runners crossing a specific location at a specific range of time can be searched. This way the tool helps in quickly spotting multiple runners running in groups, thereby reducing the time to perform the sequential search. The tool also has the feature to sort over different fields such as name, bibId, and time values. This way, we can find runners in ascending order of the time they crossed a specific location. An outlook of the ICA tool is shown in Figure \ref{fig:ICATool}.

\subsection{Performance Evaluation Metrics}
In this section, we discuss the metrics used in performance evaluation for our experiments.

\begin{figure}[ht]
\begin{center}
   \includegraphics[width=\linewidth]{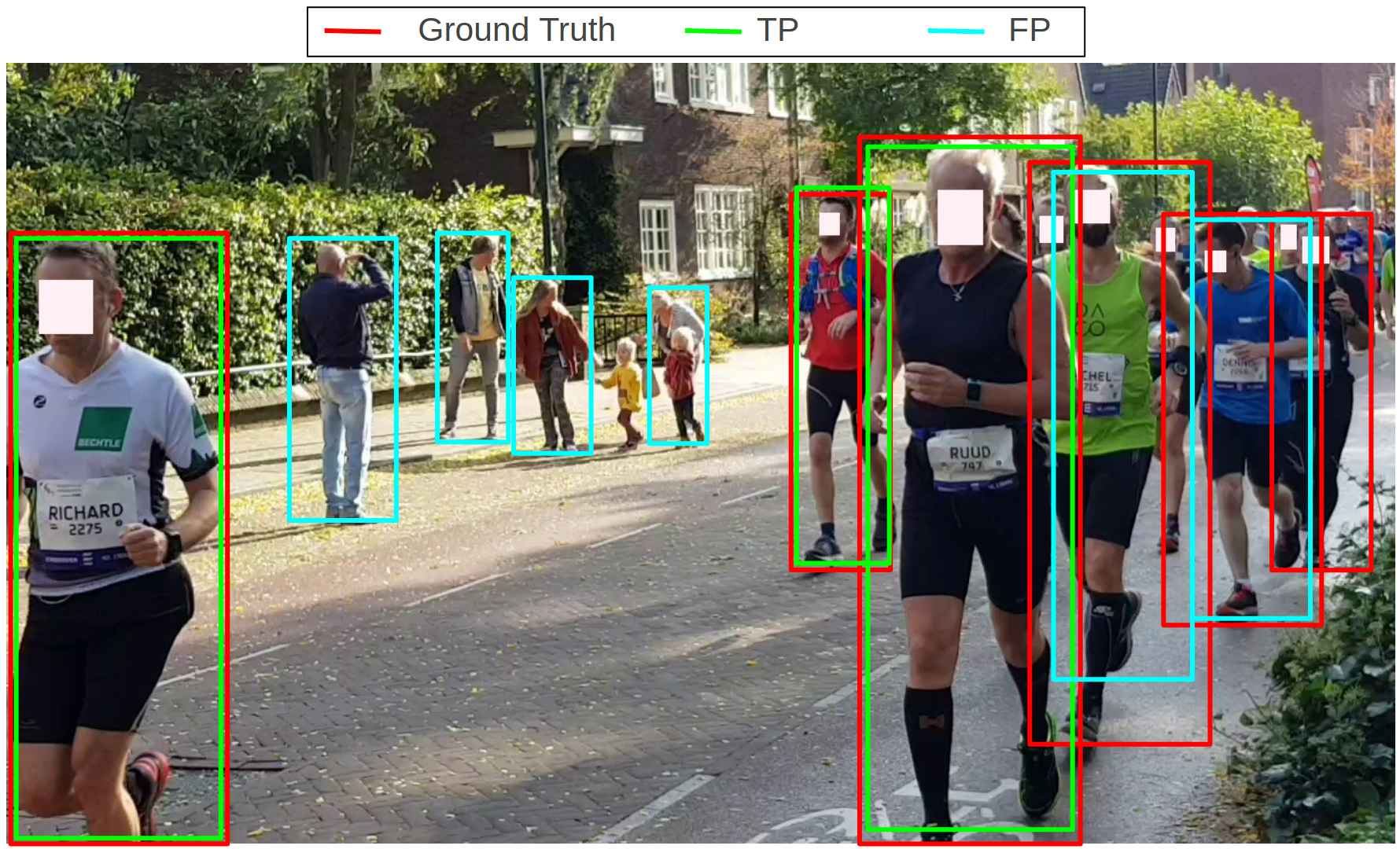}
\end{center}
   \caption{\textbf{An example of True positive (TP), False positive (FP), and False negative (FN) case.} The boxes in \textit{red} color represents the ground truth. whereas boxes in \textit{green} are TPs and in \textit{blue} are FPs. Runners that didn't get detect are FNs, represented by only \textit{red} box.}
\label{fig:tpfp}
\end{figure}

\textbf{Terminology:}
\\
\textit{True Positive (TP).} True positive is when the positive class is correctly predicted. Here, a runner is referred to as the positive class.

\textit{False Positive (FP).} False positive is when the positive class is incorrectly predicted. Here, the incorrect detection of a runner or detection of a non-runner is referred to as the FP.

\textit{False Negative (FN).} False negative is when the positive class is present but not predicted. Here, FN is when if a runner is present in the frame and is not detected.

An example is shown \ref{fig:tpfp}, explaining the cases of TPs, FPs, and FNs.

\textbf{Frame-wise Metrics.} As the video annotation is done by extracting the frames and annotating them individually, so we need a frame-wise metric to evaluate the annotation accuracy at the frame level. To determine the accuracy of the bounding box annotation,  Intersection Over Union (IoU) is used as the metric. This is given by Equation \ref{eq:IoU}

\begin{equation}
    \text{IoU} = 
\begin{cases}
    0, \hspace{30mm} \textit{if overlap}< 0\\\\
    \frac{\text{\normalsize{Area of Overlap}}}{\text{\normalsize{Area of Union}}},    \textit{\hspace{10mm} otherwise.}\\
\end{cases}
\label{eq:IoU}
\end{equation}

We define TP as given by Equation \ref{eq:tpfp}. If IoU is atleast 0.8, it's TP, else it's FP. If a runner remains undetected then it's a FN.

\begin{equation}
    \text{if},
\begin{cases}
    \text{IoU} \geqslant 0.8,  \rightarrow  \textit{TP}\\\\
    0 \leqslant \text{IoU} < 0.8,  \rightarrow  \textit{FP}\\\\
    \text{No detection},  \rightarrow   \textit{FN.}\\
\end{cases}
\label{eq:tpfp}
\end{equation}

\textbf{Video-wise Metrics.} We use Precision, Recall, and F1-Score to measure the performance of our methods on a given video sample.
Precision is defined as,

\begin{equation}
    \text{Precision \textit{(p)}} = \frac{ \text{\# TP}}{\text{\# TP} +  \text{\# FP}},
    \label{eq:precision}
\end{equation}

Recall is defined as,
\begin{equation}
    \text{Recall \textit{(r)}} = \frac{\text{\# TP}}{\text{\# TP} + \text{\# FN}},
    \label{eq:recall}
\end{equation}

F1-Score is defined as,
\begin{equation}
    \text{F1-Score} = \frac{\text{2\textit{p}\textit{r}}}{\text{\textit{p} + \textit{r}}}.
    \label{eq:f1-score}
\end{equation}

\textbf{Workload Estimation.} The total workload consists of four kinds of manual operations:

\begin{enumerate}
    \itemsep0em
    \item Removal of False Positives (Detection of non-runners)
    \item Addition of False Negatives (Missed detection of runners)
    \item Addition of labels
    \item Adjustment of boxes (Improper detection of runners)
\end{enumerate}

Total workload = (Remove FP) + (Add FN) + (Add Labels) + (Adjust detections)
\\
And therefore, total annotation time is given by Equation \ref{eq:totalannotationtime},

\begin{equation}
    \text{t}_{\text{annotation}} = \text{t}_{\text{removal}} + \text{t}_{\text{addition}} + \text{t}_{\text{adjustment}}.
\label{eq:totalannotationtime}
\end{equation}

As we have to ensure that all the runners are identified across different cameras, we use percentage of unidentified runners(UR) as the evaluation metric given by \ref{eq:unidentifiedRunnersPercent},

\begin{equation}
    \text{UR \%} = \frac{\text{\# Total runners - \# Identified Runners}}{\text{\# Total runners}} \times 100.
\label{eq:unidentifiedRunnersPercent}
\end{equation}

%------------------------------------------------------------------------

\section{Experiments}
In this section, we will discuss different experiments  corresponding to different problems and analyze the results.

\subsection*{How the overall cost of annotation can be reduced in terms of time \& budget?}

In this section, we will discuss different ways with which we can reduce the cost of annotation. 

The budget of the annotation is directly proportional to the time of annotation. Therefore, we define the cost of annotation as the total time taken to annotate the dataset. Annotation time highly depends on the type of annotation we need to perform. In our case, video annotation is performed by extracting the frames. So, there are two ways in which we can reduce the time of annotation: i) \textit{Frame level annotation without bounding box annotation. ii) By intelligently reducing the number of frames to annotate.}

\textbf{Setup.} Due to the massive amount of data, it is hard to perform evaluations on the entire dataset. So, we used four sample videos for the experiments. Each video is of 95 sec duration which is the mean duration of videos in the complete dataset. The videos are randomly selected from 4 different locations, such that they are the true representative of the entire dataset, covering various scenarios as mentioned in Figure \ref{fig:scenarios}. Following are the four selected locations with location number 3, 15, 38 and 41 and scores  10, 20, 20 and 23 respectively.

\subsection{Exp 1: Frame level annotation}

\begin{figure}[ht]
                \centering
                \includegraphics[width=\linewidth]{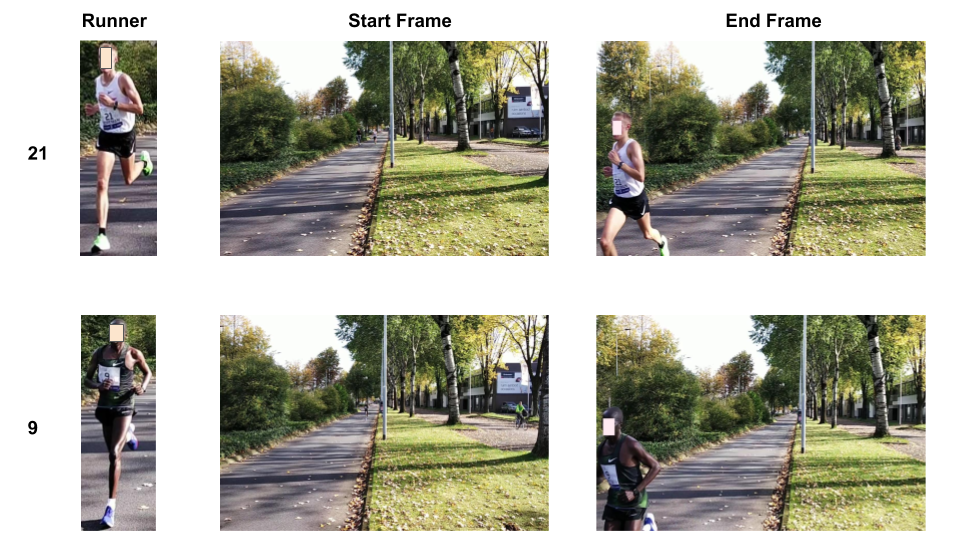}
                \caption{\textbf{An example of frame-level annotation.} The annotation is done with the target runner appearing within a range of frames. The frame range is defined as the start frame where the runner appears for the first time and end frame where a runner exists in the camera view.}
                \label{fig:frameLevelAnnotation}
\end{figure}

\begin{figure}[ht]
    \centering
    \includegraphics[width=\linewidth]{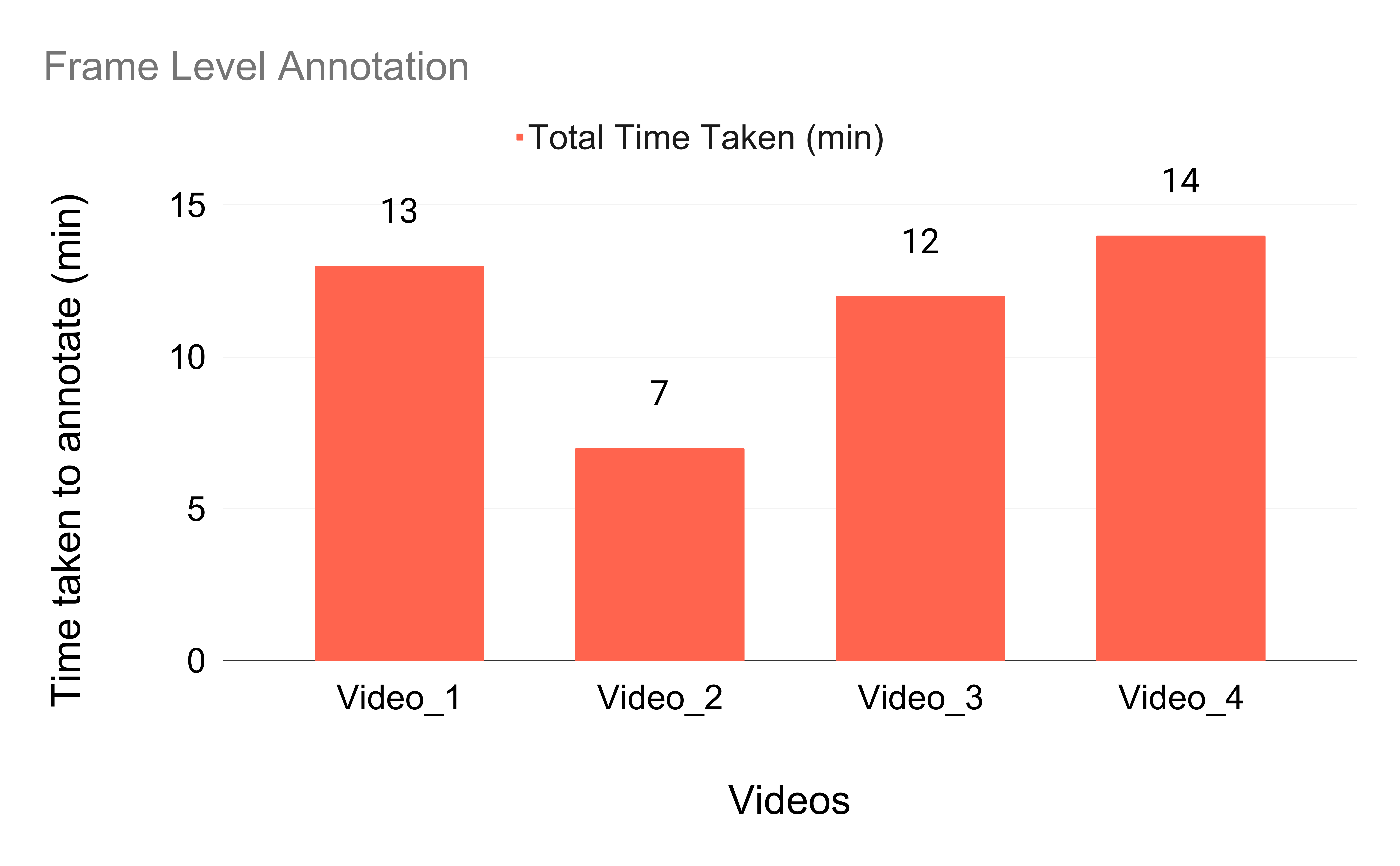}
    \caption{\textbf{Results of Frame Level Annotation.} On average, it takes around 12 min to annotate a video of 95 sec duration (mean duration).}
    \label{fig:frameLevelResult}
\end{figure}

There are different levels of details to annotate an object. The cheapest method is video-level and frame-level annotations. In frame-level annotation, video frames are manually analyzed and visualized. A range of frames is defined for each runner where they are recognizable by the human eye. A range is defined as the start frame and end frame for each individual runner in a video. \textit{Start frame} is the frame number where the runner is recognizable for the first time in the video, whereas \textit{End frame} is the frame number where a runner exits the camera's field of view. For example, if a runner 'X' is recognizable for the first time in frame number 'n1', and leaves the camera view at frame number 'n2'. Then for runner 'X', we annotate the frame 'n1' as the \textit{Start frame} and 'n2' as the \textit{End frame}. An example is shown in Figure \ref{fig:frameLevelAnnotation},

We evaluated the method on the four sample videos. The result of the experiment is shown in Figure \ref{fig:frameLevelResult}. On average it takes around 12 min to annotate a video of mean duration.
% As shown in Table \ref{table:frameLevelAnnotation}, 

There are in total 3,268 videos in the dataset. Then,
\vspace{0.2cm}

Time to annotate all videos = $12 \times 3268 = 39,216$ min.

\textbf{Analysis.}
Frame level annotation takes less time to annotate. However, the average time of annotation may vary depending on the number of runners in the video. That is, if there are hundreds of runners running together, then most of the runners would be occluded and therefore, it will take more time to spot and annotate the runner.

Overall, frame-level annotation is an easy and naive way to annotate which helps to reduce the cost of annotation. However, in this type of annotation, the use cases of the dataset will be limited, as we don't have pixel-level annotation.

\subsection{Exp 2: Analysing the effect of FPS on annotation time}
The main question behind this experiment is to investigate whether we can obtain the same amount of information from the videos if we extract the frames at some different extraction rates. If we can reduce the frame rate per second, then the total number of frames will be reduced, leading to the reduced time and effort of annotation.

\textbf{Setup.}
We evaluate the method on 4 sample videos. The videos are recorded at \textit{30fps}. The experiment is performed at five different \textit{fps} values, $n = \{1, 2, 5, 10, 30\}$, where n = frames per second. In each case, every $(30/n)\textsuperscript{th}$ frame is extracted. For example, if $n= 2$, then we take every 15th frame and remove the others, or in other words, there are 2 frames in 1 second. We used LabelImg~\cite{labelimg}, for manually creating the bounding boxes. A runner's bibId is used as the label.

\begin{figure}[ht]
\begin{center}
     \subfloat[Annotation Time Comparison]{{\includegraphics[width=\linewidth]{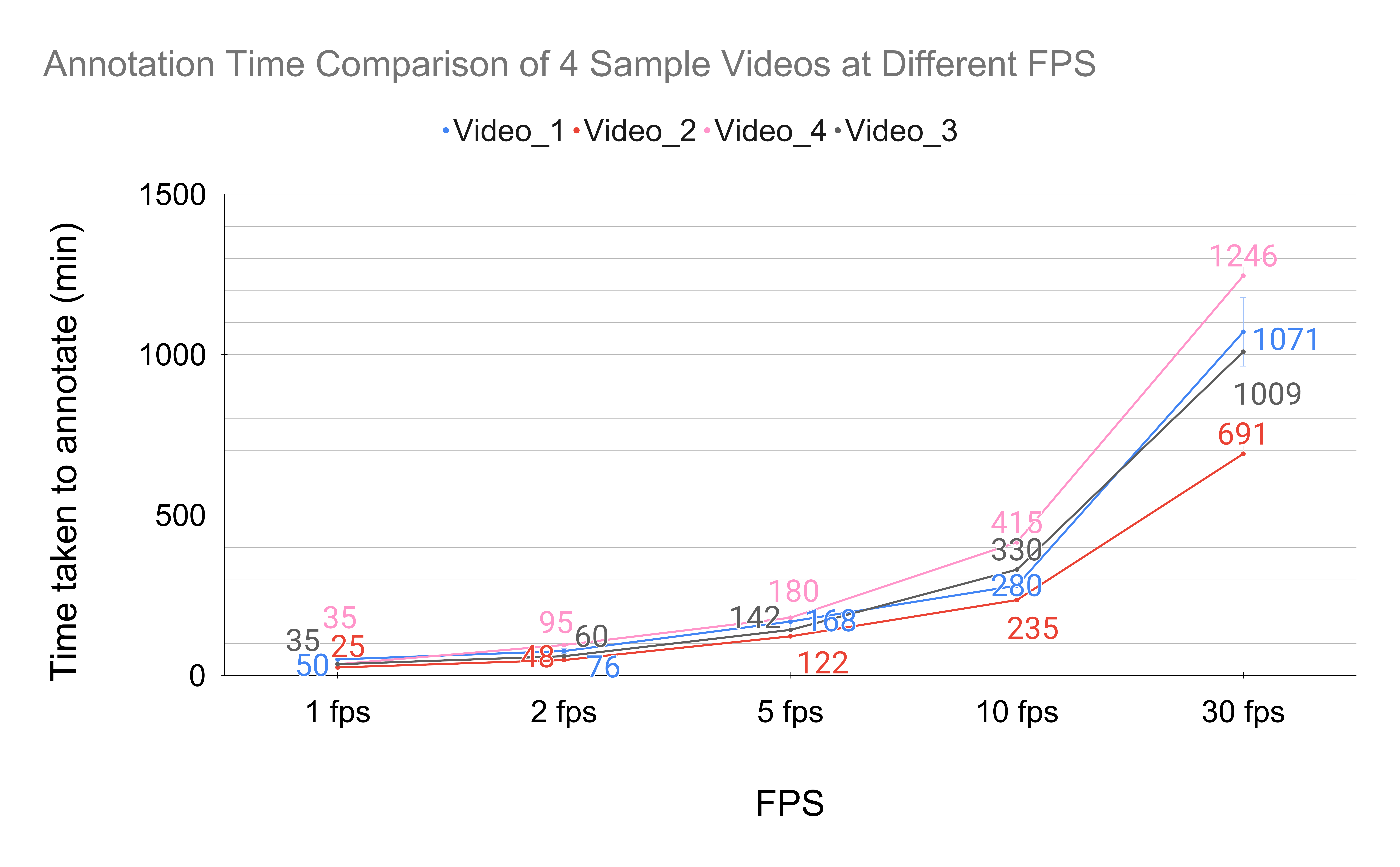} }}%
     \hfill
     \subfloat[Analysing the average annotation time and percentage of unidentified runners]{{\includegraphics[width=\linewidth]{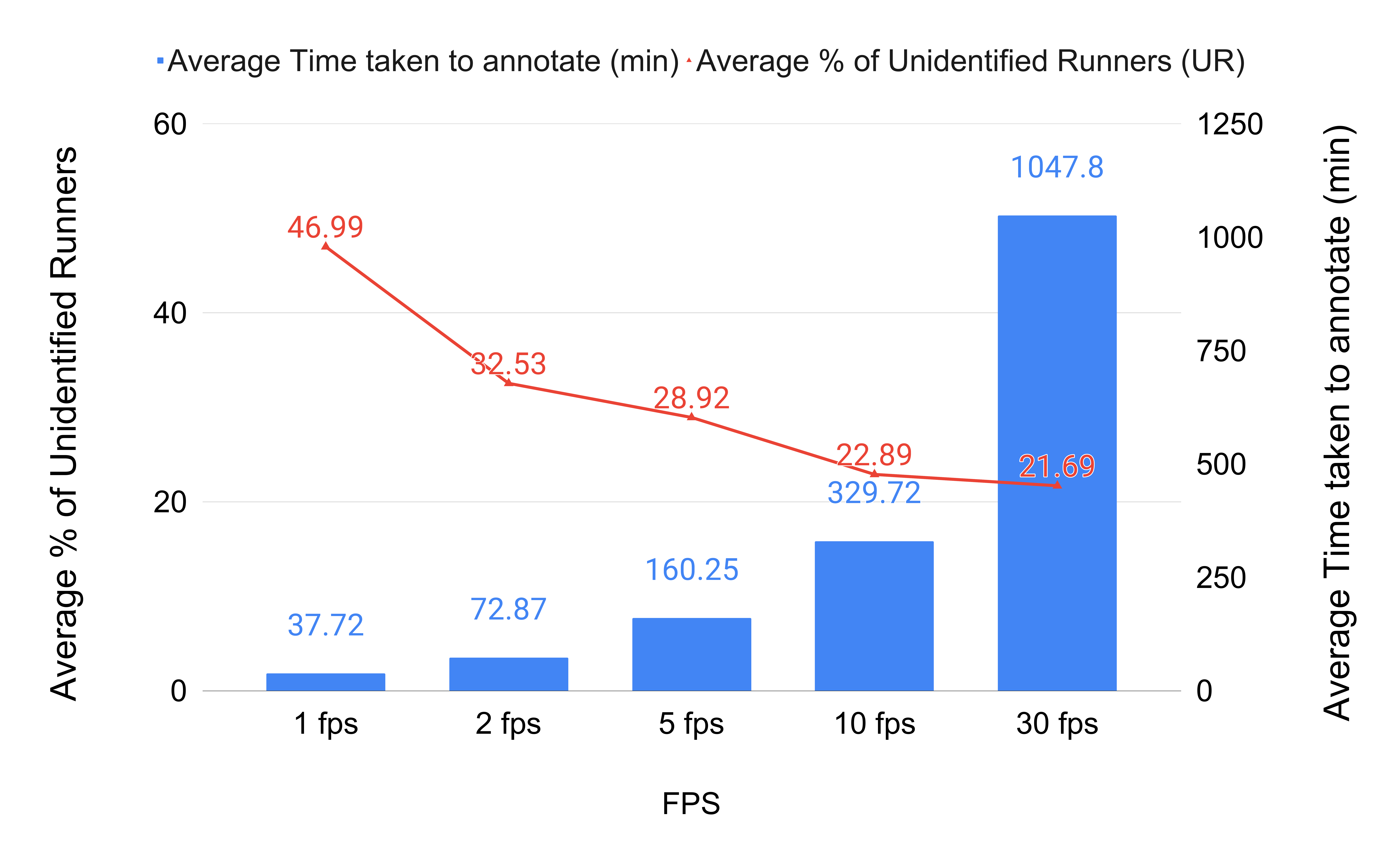} }}%
    \hfill
\end{center}
   \caption{\textbf{Results of the FPS experiment are shown in figure.} (a) Comparison of time taken to annotate four sample videos at different fps rates is shown in the figure. It is clear that annotation time increases with the increase in the number of frames per second. (b) In the figure, as the frame extraction rate increases, more number of runners are identified. But the slope of the curve doesn't change much after 5fps. Also, the annotation time increases with the increase in fps, as the number of frames to be annotated also increases.}
\label{fig:timeComparisonFPSAnalysis}
\end{figure}

\textbf{Analysis.}
Results in Fig\ref{fig:timeComparisonFPSAnalysis}(a) shows that it takes more time to annotate a video if it's recorded at a higher frame rate. Also, from Fig\ref{fig:timeComparisonFPSAnalysis}(b), it is clear that as the fps increases, the number of unidentified number decreases. Also, there is very little change in the slope of the curve after 5fps. Most of the runners' identities visible in the 30fps case are also visible in the case of 5 fps videos. On an average, we are able to identify $91.8\%$ of the identifiable runners in 5fps videos. We conclude that it's better to annotate at a lower fps rate (in this case 5 fps), as it takes a lesser amount of time to annotate, and at the same time derives most of the information from the data.

\subsection*{How can the efficient bounding-box annotation be done?}
Another problem in the annotation is to generate accurate ground truth. The performance of most of the machine learning model highly depends on the accuracy of the ground truth labels. So, it becomes important to generate tight bounding boxes around the runners for the ground truth labels.

There are several ways to do the bounding box annotation. We have analyzed and compared three different approaches namely \textit{object detector, object tracking and box interpolation,} alongside the baseline method. For all the experiments, we use annotation time as a common unit to compare the efficiency of one approach with other approaches.

% \begin{figure*}
% \begin{center}

%     {{\includegraphics[width=0.32\textwidth]{images/Experiments/BoundingBoxAnnotation/OD_FPFN.pdf} }}%
%     \hfill
%     {{\includegraphics[width=0.32\textwidth]{images/Experiments/BoundingBoxAnnotation/MOT_FPFN.pdf} }}%
%     \hfill
%     {{\includegraphics[width=0.32\textwidth]{images/Experiments/BoundingBoxAnnotation/Interpolation_FPFN.pdf} }}%
%     \hfill
    
%      \subfloat[Object Detector Method Results]{{\includegraphics[width=0.32\textwidth]{images/Experiments/BoundingBoxAnnotation/OD_TimeAnalysis.pdf} }}%
%     \hfill
%     \subfloat[Multi-Object Tracking Method Results]{{\includegraphics[width=0.32\textwidth]{images/Experiments/BoundingBoxAnnotation/MOT_TimeAnalysis.pdf} }}%
%     \hfill
%     \subfloat[Interpolation Method Results]{{\includegraphics[width=0.32\textwidth]{images/Experiments/BoundingBoxAnnotation/Interpolation_TimeAnalysis.pdf} }}%
% \end{center}
%   \caption{\textbf{Comparison of results of different methods of bounding box generation.} (a) Time taken in the annotation method using Object Detector is more than the baseline method.  This is due to the presence of too many FPs in the form of non-runners, which needs to be removed manually. (b, c) Time taken in annotation using MOT and Interpolation is less than the baseline method. It can be observed that the number of FPs are very less. Also, the number of TPs is high, resulting in a reduced time of annotation.}
% \label{fig:boundingBoxExperimentsResults}
% \end{figure*}

\subsection{Exp 3: Baseline: Manually generating the bounding boxes}  
\begin{figure}[ht]
\begin{center}
   \includegraphics[width=\linewidth]{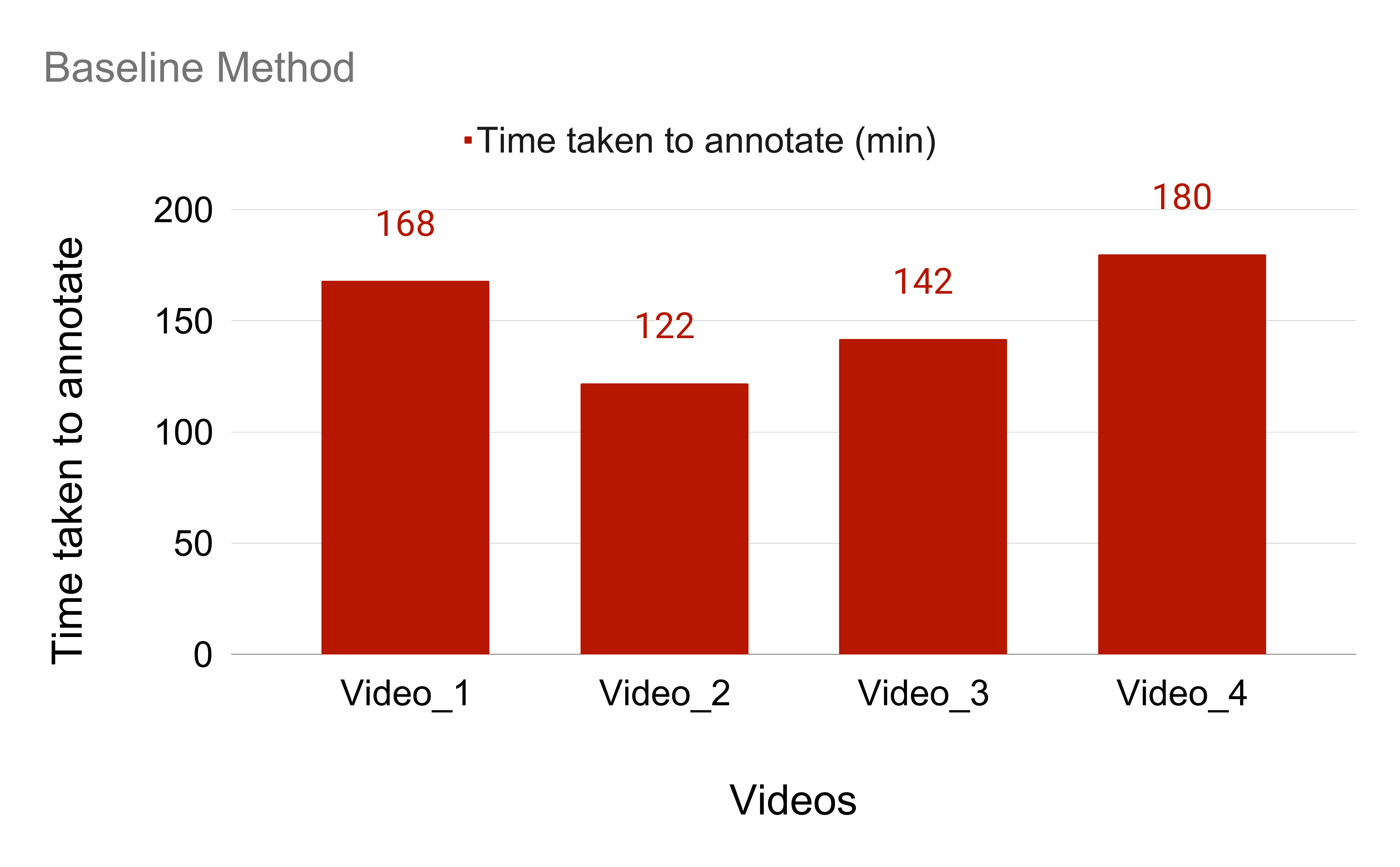}
\end{center}
   \caption{\textbf{Results of the baseline method.} Manual bounding box annotation is performed on four sample videos. On average, it takes 153 minutes to annotate a video of mean duration.}
\label{fig:baselineMethodBbox}
\end{figure}

We consider the manual annotation performed at 5 fps extraction rate in the FPS experiment, as our main baseline method. Results of the baseline method is shown in Figure \ref{fig:baselineMethodBbox}. On average, it takes around \textit{153 min} to generate bounding boxes in a video of 95 sec duration (mean duration).

\begin{figure*}
\begin{center}

    \subfloat[Object Detector Method Results]{{\includegraphics[width=0.99\textwidth]{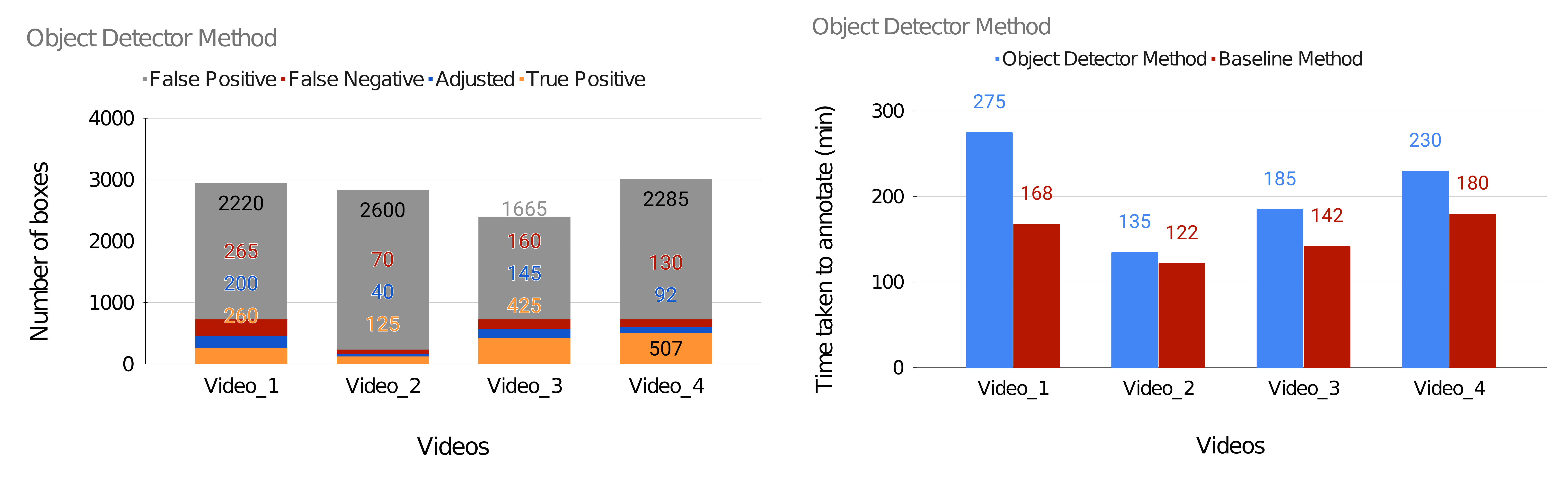} }}%
    \hfill
    
    \subfloat[Multi-Object Tracking Method Results]{{\includegraphics[width=0.99\textwidth]{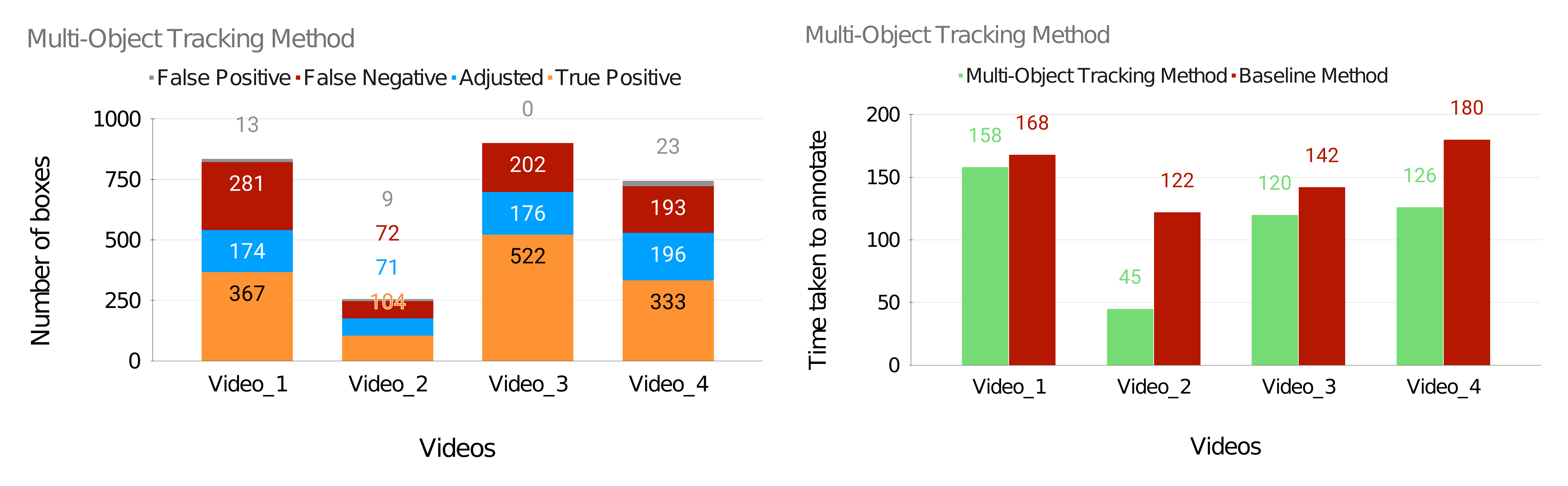} }}%
    \hfill
    
    \subfloat[Interpolation Method Results]{{\includegraphics[width=0.99\textwidth]{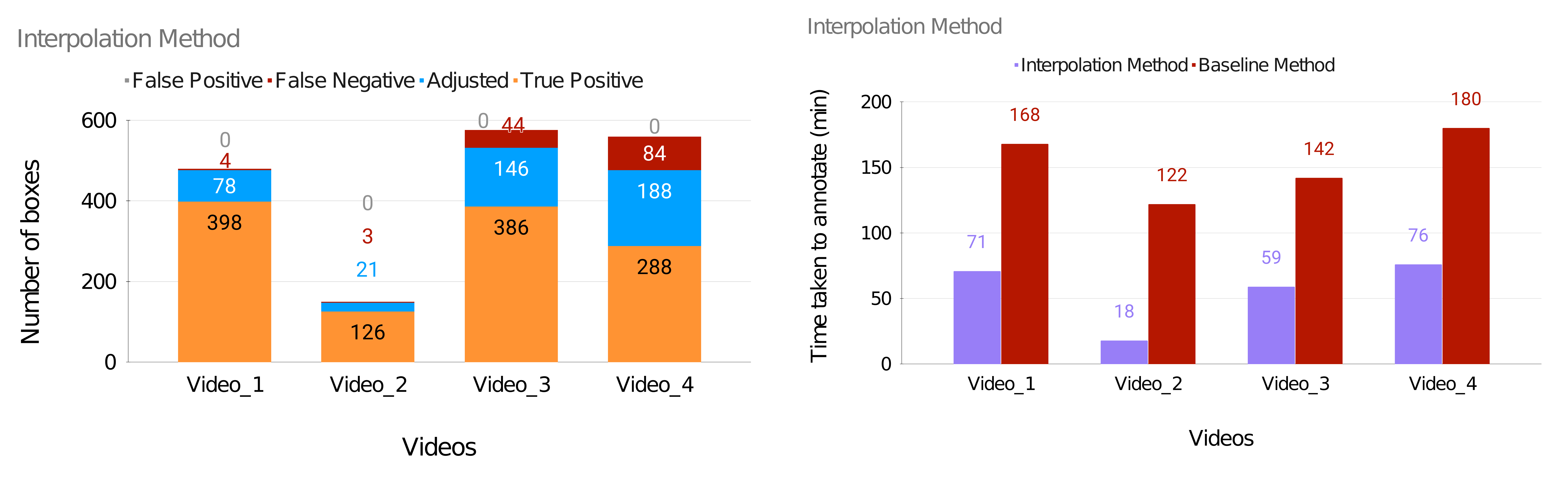} }}%
\end{center}
  \caption{\textbf{Comparison of results of different methods of bounding box generation.} (a) Time taken in the annotation method using Object Detector is more than the baseline method.  This is due to the presence of too many FPs in the form of non-runners, which needs to be removed manually. (b, c) Time taken in annotation using MOT and Interpolation is less than the baseline method. It can be observed that the number of FPs are very less. Also, the number of TPs is high, resulting in a reduced time of annotation.}
\label{fig:boundingBoxExperimentsResults}
\end{figure*}

\begin{figure*}
\begin{center}
    
     \subfloat[Precision, Recall and F1-Score comparison ]{{\includegraphics[width=0.49\linewidth]{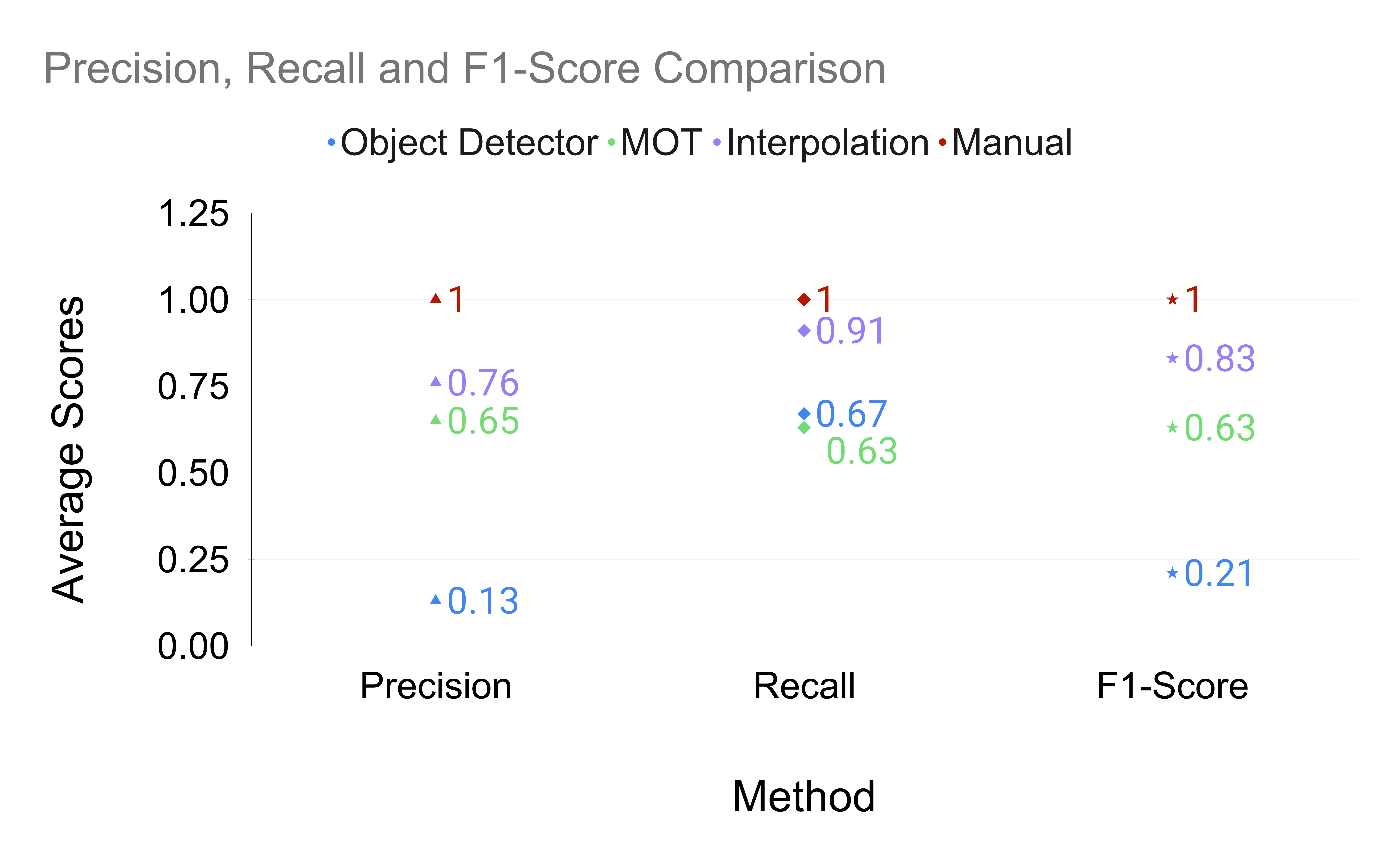} }}%
    \hfill
    \subfloat[Annotation time comparison of different methods]{{\includegraphics[width=0.49\linewidth]{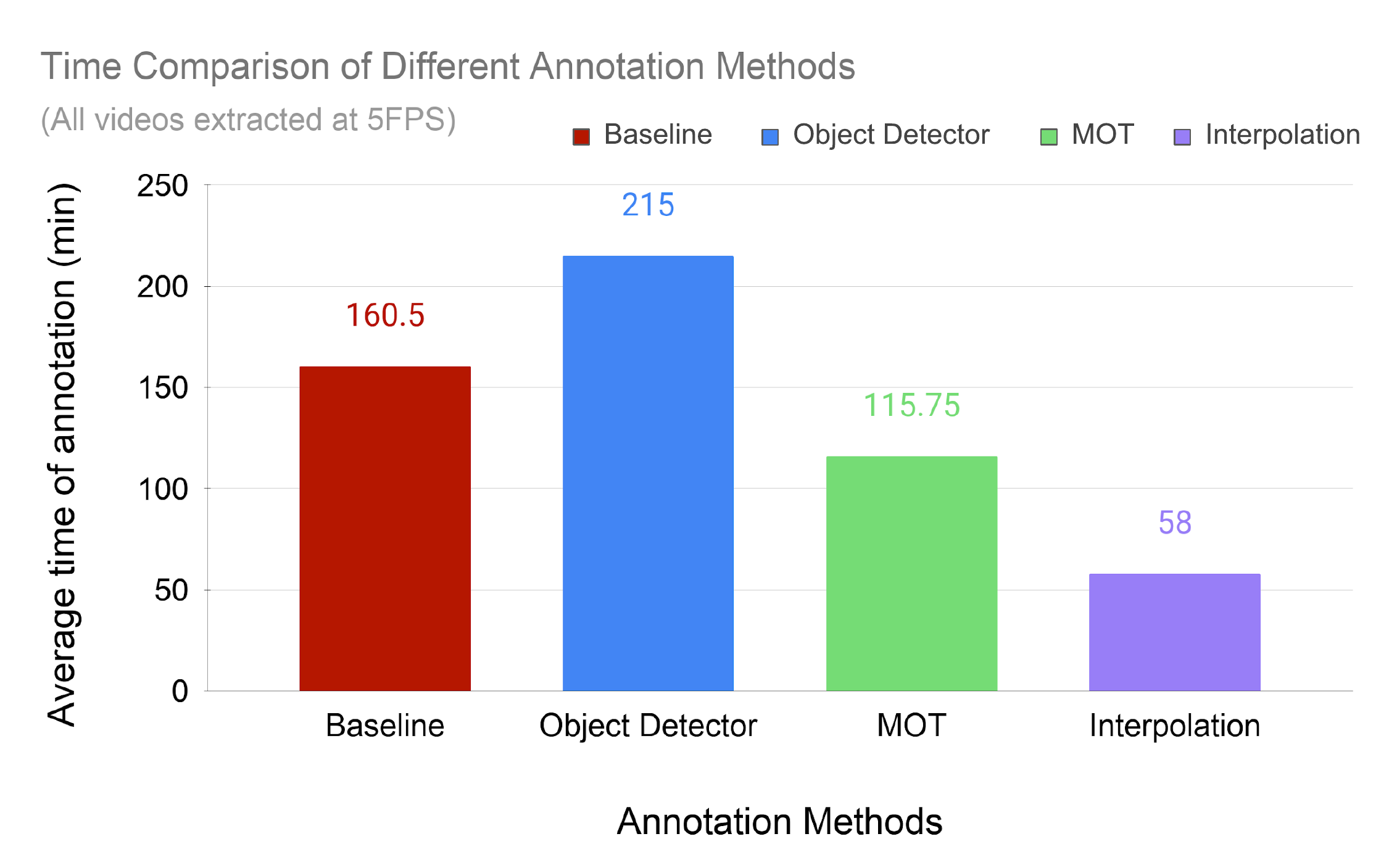} }}%
    
\end{center}
   \caption{\textbf{Accuracy and Time Comparison of different annotation methods.} (a) It is clear that the annotation using interpolation has maximum precision and recall values. It is because the number of false positives and false negatives in the interpolation method is less than any other method. (b) The annotation time using the interpolation method is least compared to any other method.}
\label{fig:annotationTimeComparison}
\end{figure*}

\subsection{Exp 4: Faster bounding box annotation using Object Detector}
Our interest is to validate the hypothesis if the automatic generation of bounding boxes using object detectors could save the time and effort of annotation.

We used the same four sample videos for experimentation as used in the baseline method. The frames are extracted at 5FPS rate. Next, we feed the frame into the deep learning based object detector models for the generation of bounding boxes. Any SOTA object detector can be used. We used YoloV3~\cite{yolov3} pre-trained model, trained on the COCO dataset. As our main target is to predict a runner, therefore we removed all the classes except the person class. This way we prevent the generation of unwanted bounding boxes.

\textbf{Manual correction of inferred predictions.} The incorrectly predicted boxes are adjusted or removed by a human annotator. The annotator has to go through all the proposed boxes and check if the predictions are correct or need an adjustment. If IoU is greater than 0.8, then it is TP, else it is an FP. All the FPs over the runners are adjusted, whereas the ones including the non-runners are removed manually. A runner's bibId is used as the label and is manually added for each prediction by the annotator. The total time of annotation is calculated using Equation \ref{eq:totalannotationtime}.

% \begin{figure}[ht]
% \begin{center}
%     \subfloat[Analyzing the number of TPs, FPs, FNs and number of boxes adjusted]{{\includegraphics[width=\linewidth]{images/Experiments/BoundingBoxAnnotation/OD_FPFN.png} }}%
%     \hfill
%     \subfloat[Annotation time comparison of the object detector method with the baseline method]{{\includegraphics[width=\linewidth]{images/Experiments/BoundingBoxAnnotation/OD_TimeAnalysis.png} }}%
% \end{center}
%   \caption{Results of the Object Detector method. Time taken in semi-automatic annotation method is more than the baseline method. This is due to the presence of too many FPs in the form of non-runners, which needs to be removed manually.}
% \label{fig:od_results}
% \end{figure}

\textbf{Analysis.} 
The results of the experiment are shown in Figure \ref{fig:boundingBoxExperimentsResults}(a). It can be seen in Fig.\ref{fig:boundingBoxExperimentsResults}(a), that the number of FPs are too much. The FPs are due to the detection of the non-runners which are present in every frame. These FPs are the unwanted detections which are unavoidable, and it needs to be removed manually from every frame. This results in an increase in the overall annotation time. Hence, the total time in the case of the semi-automatic annotation using an object detector is more than the baseline method. The average precision, recall and F1-score is shown in Table \ref{table: od_evaluationMetrics}. The lower precision value is due to a higher percentage of false positives.

\begin{table}[ht]
\begin{center}
\begin{tabular}{c c c}
\toprule
\textbf{Precision} &   \textbf{Recall} & \textbf{F1-Score} \\ 
\midrule
$0.13$ &   $0.67$ & $0.21$ \\ 
\bottomrule
\end{tabular}
\caption{\textbf{Average Precision, Recall and F1-Score of Object Detector Method.} The lower precision value is due to a higher percentage of false positives due to the presence of non-runners in most of the frames.}
\label{table: od_evaluationMetrics}
\end{center}
\end{table}

We conclude that using an object detector to automatically generate the bounding boxes saves time in the case when only the object of interest is present in the frame. In the case of our dataset, the method completely fails, due to the presence of too many false positives, as a result, it takes more time than the baseline method.

\subsection{Exp 5: Faster bounding box annotation using Multi-Object Tracking}

\begin{figure*}
\begin{center}
    \subfloat[Trajectory annotation]{{\includegraphics[width=0.3\textwidth, height=0.18\linewidth]{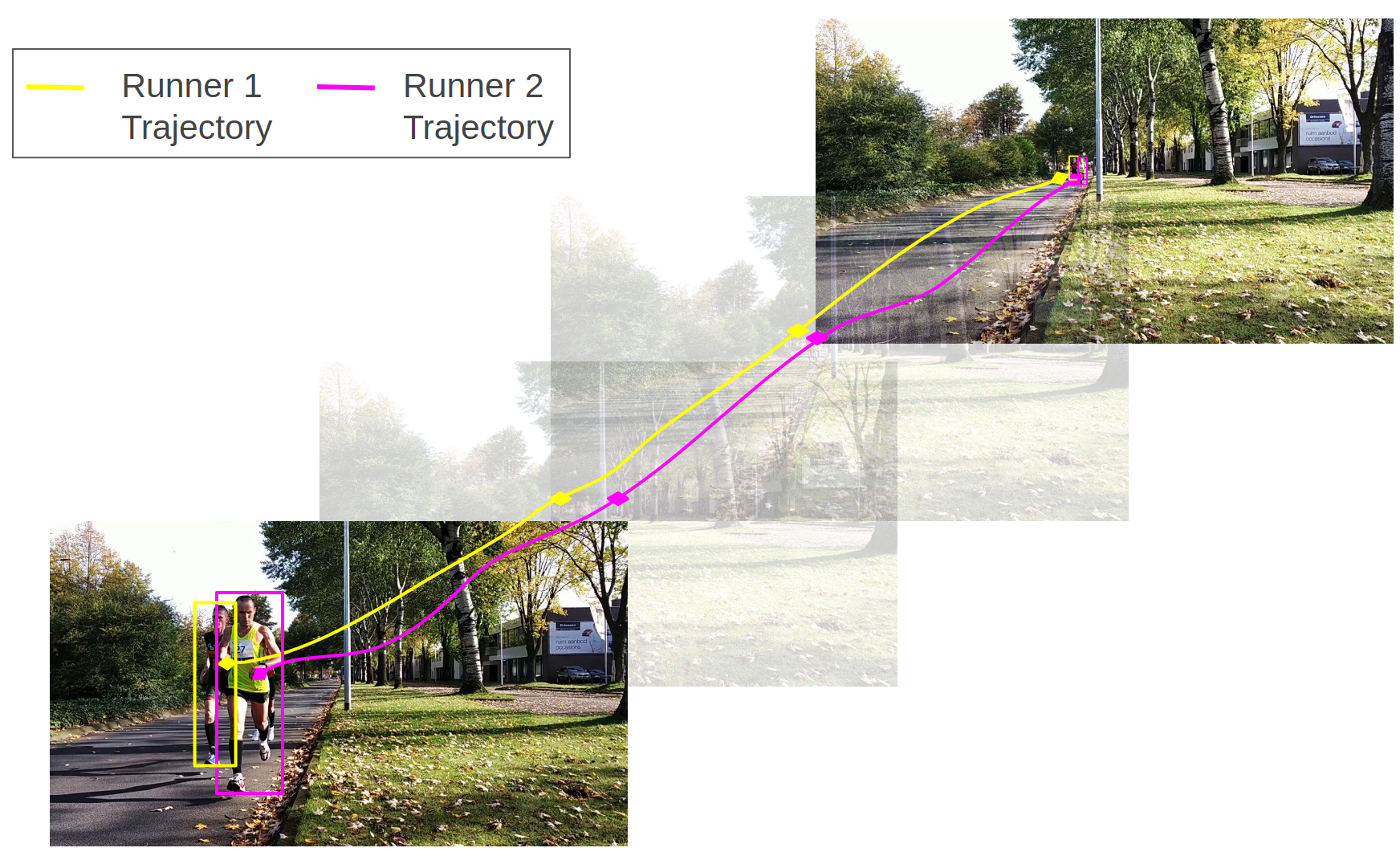} }}%
    \hfill
    \subfloat[Linking Trajectory and Box annotations]{{\includegraphics[width=0.65\textwidth]{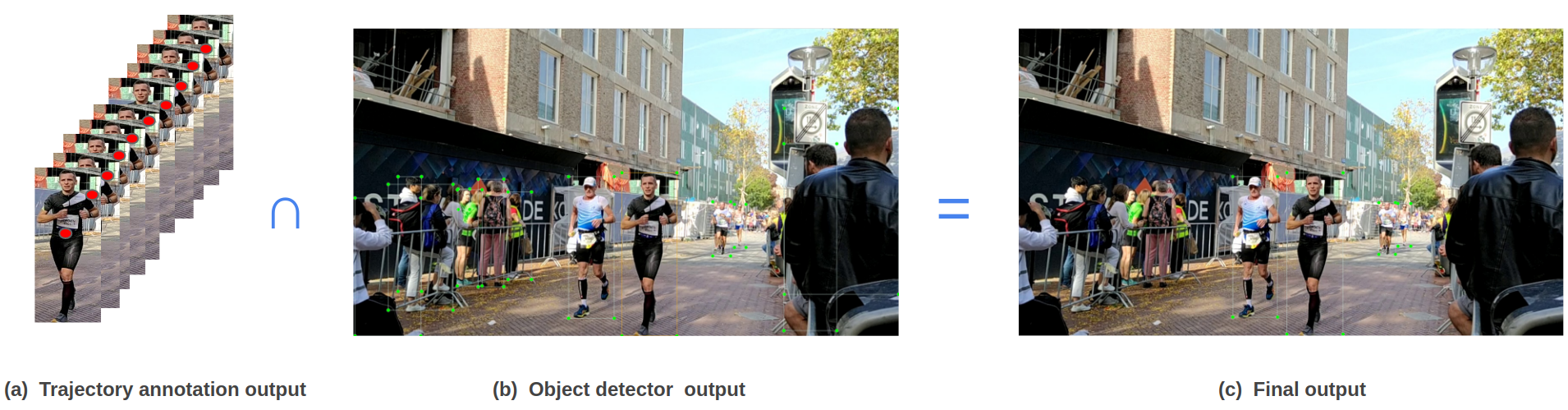} }}%
\end{center}
   \caption{\textbf{The figure shows the annotation procedure of the MOT method.} (a) The trajectory of runners is followed using the mouse cursor. The colored lines show the mouse movement during the path supervision. (b) The intersection of the trajectory annotation and the object detector detection outputs the final annotation. It helps in removing the false positives detected in the object detector.}
\label{fig:mot_procedure}
\end{figure*}

A typical annotation pipeline would involve an annotator watching the videos or images while doing annotations. So, our main idea is to verify if we can make use of this watching time of video by efficiently turning it into annotation time. Also, in this way, we can overcome the problems of object detector experiment, by avoiding the unwanted predictions of non-runners.

The idea is based on the paper \textit{Pathtrack}~\cite{pathtrack}, where the authors use path supervision to generate dense box trajectory annotations for Multi-Object Tracking (MOT) datasets.

\textbf{Setup.} For experiments, we use the same set of four sample videos. Firstly, the frame rate of videos is decreased from 30fps to 5fps. Then we use the Pathtrack Tool~\cite{pathtrack}, to annotate these videos.

The tool has an interface to play a video. The user can control the playback speed by speeding up or down, as per the requirement. For annotation of runners, firstly a segment of the video where the target runner is visible is played once, to observe the trajectory of the runner. Then, the user rewinds the video back to the point where the runner is visible for the first time. The user then changes the playback speed of the video and starts following the runner's trajectory by hovering the mouse cursor near the center of the runner. In the final trajectory, the annotator provides three bounding boxes for the first and last appearance of the runner and one in between the two. 

\textbf{Steps.} The annotation using path supervision is performed mainly in two stages: 

\textit{Trajectory annotation with path supervision.} Annotations using path supervision is efficient \& intuitive and is obtained by watching each runner independently and tracking it using a mouse cursor. Annotation path of a runner \textit{r} consists of an (x, y)-coordinate point $\text{p}_{r}\text{(f)}$ that lies inside it's location boundaries at frame id $\text{f}$.

\textit{Bounding box generation.} The trajectory annotation will provide the coordinates of a runner in every frame. Now, we need to generate bounding boxes over the runners. We use an automatic way to predict the bounding boxes by using deep learning-based object detector models. We used YoloV3~\cite{yolov3} pre-trained model, trained on the COCO dataset~\cite{coco}, to predict boxes only for \textit{person} class. The object detector will predict the boxes, both over runners and non-runners.

\textit{Linking Trajectory and Box annotations.} The main task now is to remove the unwanted FPs and generate box over the trajectories of runners. Given the set of path annotations $\text{p}_{r}\text{(f)}$ and object detection D, the intersection of the two annotations will remove the FPs. In Equation \ref{eq:final_detections_mot}, in a given frame f, if the detection D$_{i}$ contains at least one point  $\text{p}_{r}\text{(f)}$, such that point $\text{p}_{r}\text{(f)}$ lies inside or over D$_{i}$, then the detection survives, else it is eliminated.

If, detection $D_{i}$ has coordinates $x_{min}, y_{min}, $ and $x_{max}, y_{max}$, then

\begin{equation}
\text{$D_i$} =
    \begin{cases}
    \text{Accepted},& \text{if } \text{x}_{min} \leqslant \text{x},  \leqslant  \textit{x}_{max} \text{ \& }\\
    \text{  }  &\text{\hspace{3mm}y}_{min} \leqslant \text{y} \leqslant \text{y}_{max}\\
    \text{Rejected},              & \text{otherwise.}
\end{cases}
\label{eq:final_detections_mot}
\end{equation}

This way we get bounding boxes generated over the runners along their trajectories. Each runner's detection is then labeled manually, by assigning the runners' bibId as the label. Thereafter, the unavoidable false positives are removed and adjusted, whereas FNs are added manually by the annotator. Figure \ref{fig:mot_procedure} describes the annotation procedure using MOT method.

\textit{Workload estimation.} The total time of annotation includes mainly 5 time-consuming components: 1) Watching the video every time to check a runners' trajectory once, 2) Following each runners' trajectory using mouse hovering 3) Time to adjust and create the bounding boxes 4) Time to remove the false positives, 5) Time to add the labels for each box prediction.

% \begin{figure}[ht]
% \begin{center}
%     \subfloat[Analyzing the number of TPs, FPs, FNs and number of boxes adjusted]{{\includegraphics[width=\linewidth]{images/Experiments/BoundingBoxAnnotation/MOT_FPFN.png} }}%
%     \hfill
%     \subfloat[Annotation time comparison of the multi-object tracking method with the baseline method]{{\includegraphics[width=\linewidth]{images/Experiments/BoundingBoxAnnotation/MOT_TimeAnalysis.png} }}%
% \end{center}
%   \caption{Results of the Multi-Object Tracking method. Time taken in semi-automatic annotation method is less than the baseline method. It can be observed that the number of FPs are very less. Also, the number of TPs is high, resulting in the reduced time in annotation.}
% \label{fig:mot_results}
% \end{figure}

\textbf{Analysis.} The results of the experiments are shown in Figure \ref{fig:boundingBoxExperimentsResults}(b). It is clear from Fig \ref{fig:boundingBoxExperimentsResults}(b), the semi-automatic annotation using MOT takes lesser time compared to the baseline annotation. The method generates a quick path trajectory annotation of runners. It generates bounding boxes 1.36x times faster than the baseline method. Also, it reduces the number of false positives mainly due to the avoidance of non-runners detection, see Figure \ref{fig:boundingBoxExperimentsResults}(b). There's also an improvement in precision, recall, and F1-score, compared to the object detector experiment, see Table \ref{table: mot_evaluationMetrics}.

However, the method fails to generate boxes in case the runner is far away from the camera frame. It is due to the inaccurate prediction of the object detector. Also, it is difficult to track runners in a crowded scene, as the runner's visibility is not constant throughout the video segment because of occlusion. Due to this, the number of false negatives is high resulting in decreased recall value.

\begin{table}[ht]
\begin{center}
\begin{tabular}{c l l l}
\toprule
\textbf{Precision} &   \textbf{Recall} & \textbf{F1-Score} \\ 
\midrule
$0.65$ &   $0.63$ & $0.63$ \\ 
\bottomrule
\end{tabular}
\caption{\textbf{Average Precision, Recall and F1-Score of Multi-Object Tracking Method.} The recall value is lower because of higher number of FNs. It is difficult to track runners in crowded scenes, hence the number of FNs is high.}
\label{table: mot_evaluationMetrics}
\end{center}
\end{table}

\subsection{Exp 6: Faster bounding box annotation using Bounding Box Interpolation}
In the baseline annotation method, the annotator has to put a bounding box around every runner in every frame. As the position of the runner doesn't vary much in consecutive frames, it could be useful if boxes can be interpolated between frames. Our main hypothesis is that interpolating the boxes between keyframes can save a lot of time, as we can assume that the trajectory of the runner is mostly linear.

The interpolation technique is intuitive, efficient, easy to implement, and produces compelling results. In our approach, we annotate a sparse-set of bounding boxes known as \textit{keyframes}, and linearly interpolate between them. Keyframes are defined as the frames where the bounding boxes if created, would help in filling the boxes in intermediate frames. We used \textit{Darklabel}~\cite{darklabel} tool that uses a linear interpolation technique to propagate the boxes for the intermediate frames in between the given set of keyframes.

\begin{figure}[ht]
\begin{center}
   \includegraphics[width=\linewidth]{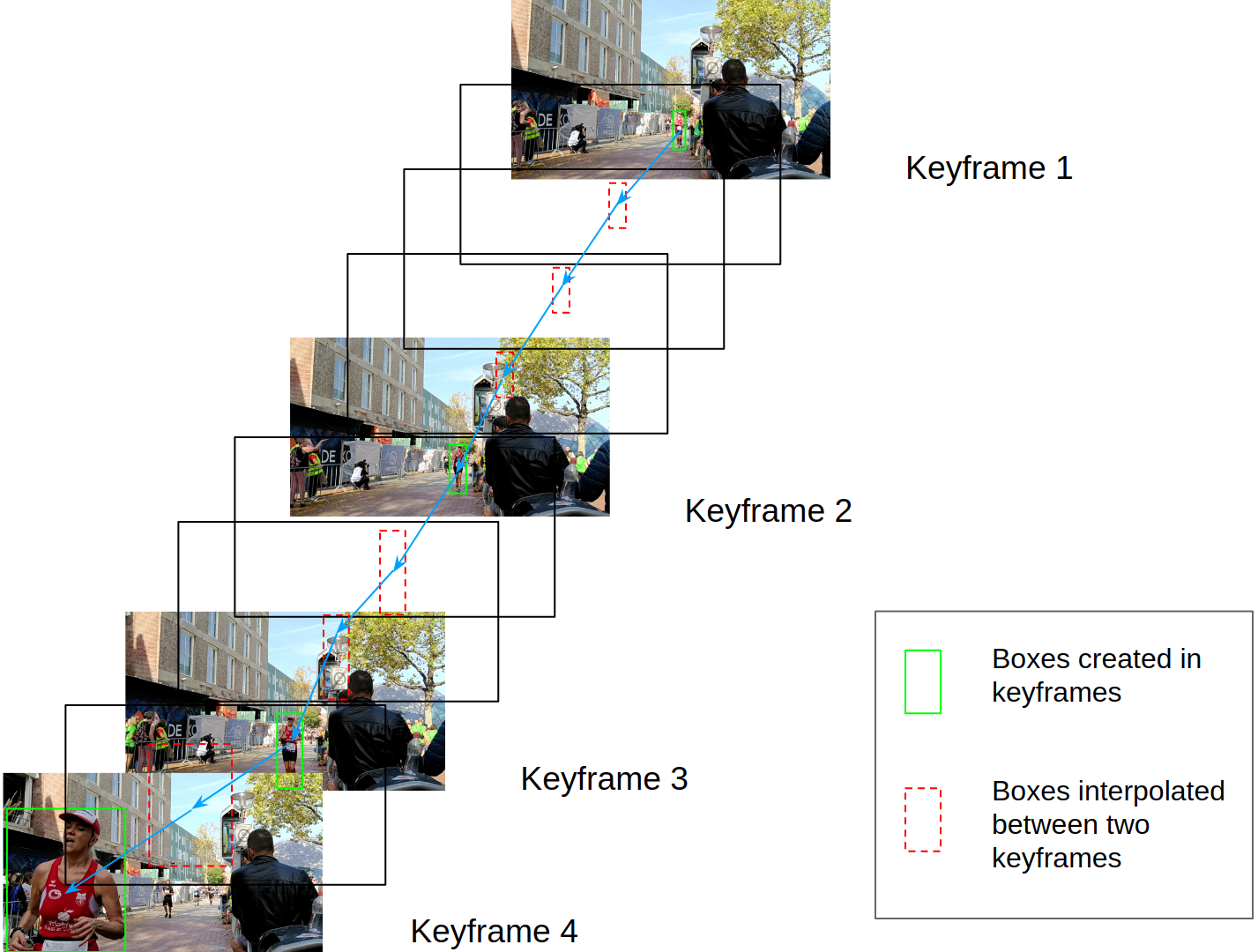}
\end{center}
   \caption{\textbf{The workflow of the interpolation method is shown in Figure.} The bounding boxes are created only in keyframes, and the rest of the boxes are linearly interpolated in the frames in between consecutive keyframes. As can be noticed, the size of the runner increase as it approaches near to the camera, and therefore the keyframe interval is decreased.}
   
\label{fig:interpolationWorkflow}
\end{figure}

\textbf{Setup.} The four sample videos are used for the experiment. Firstly, the frame rate of videos is decreased from 30fps to 5fps. We used Darklabel~\cite{darklabel} annotation tool for interpolating the boxes. The annotator selects a runner and finds his bibId. Then a frame is selected where the runner is visible for the first time in the camera. The annotator starts with putting the bounding box around the runner and labels it with it's bibId. The next four frames are skipped and the next box is created in the 5th frame. The steps are repeated until the runner comes closer to the camera. As it is more difficult to adjust a box than to create a new one. So, the keyframe interval is decreased as the runner approaches towards the camera, to avoid as many FPs. It is because the size of the runner increases as it comes closer to the camera, and therefore the boxes can't be linearly interpolated across distant keyframes. The overflow workflow of the method is explained in Figure \ref{fig:interpolationWorkflow}.

\textbf{Workload.} The total time of annotation includes the time to create boxes in keyframes, adjustment of FPs, the addition of FNs, and time to label the detections.

\textbf{Analysis.} The results of the experiment are shown in Fig.\ref{fig:boundingBoxExperimentsResults}(c). The method clearly outperforms the baseline method by a large margin. The method generates bounding boxes almost 3x times faster than the baseline method. In Fig.\ref{fig:boundingBoxExperimentsResults}(c), the number of FPs are very low, and hence the precision is very high, as shown in Table \ref{table: interpolation_evaluationMetrics}. The number of FNs is also very less compared to any previous method used, and hence the method has the highest recall value of $91\%$. The number of adjustments in Video\_3 and Video\_4 are high due to the the presence of shakiness of the camera. Due to shakiness, the runner location shifts drastically between consecutive frames. This is where linear interpolation incorrectly propagates the box, and therefore adjustment is needed.

\begin{table}[ht]
\begin{center}
\begin{tabular}{c l l l}
\toprule
\textbf{Precision} &   \textbf{Recall} & \textbf{F1-Score} \\ 
\midrule
$0.76$ &   $0.91$ & $0.83$ \\ 
\bottomrule
\end{tabular}
\caption{\textbf{Average Precision, Recall and F1-Score of Interpolation Method.} The precision  and recall values are high because of very less FPs and FNs respectively.}
\label{table: interpolation_evaluationMetrics}
\end{center}
\end{table}

\subsection*{How can Inter-Camera alignment be achieved in minimum time and effort?}

\begin{figure*}[!b]
\begin{center}
    \subfloat[Comparing the number of bibId's found and not found using three different methods of ICA]{{\includegraphics[width=0.48\textwidth]{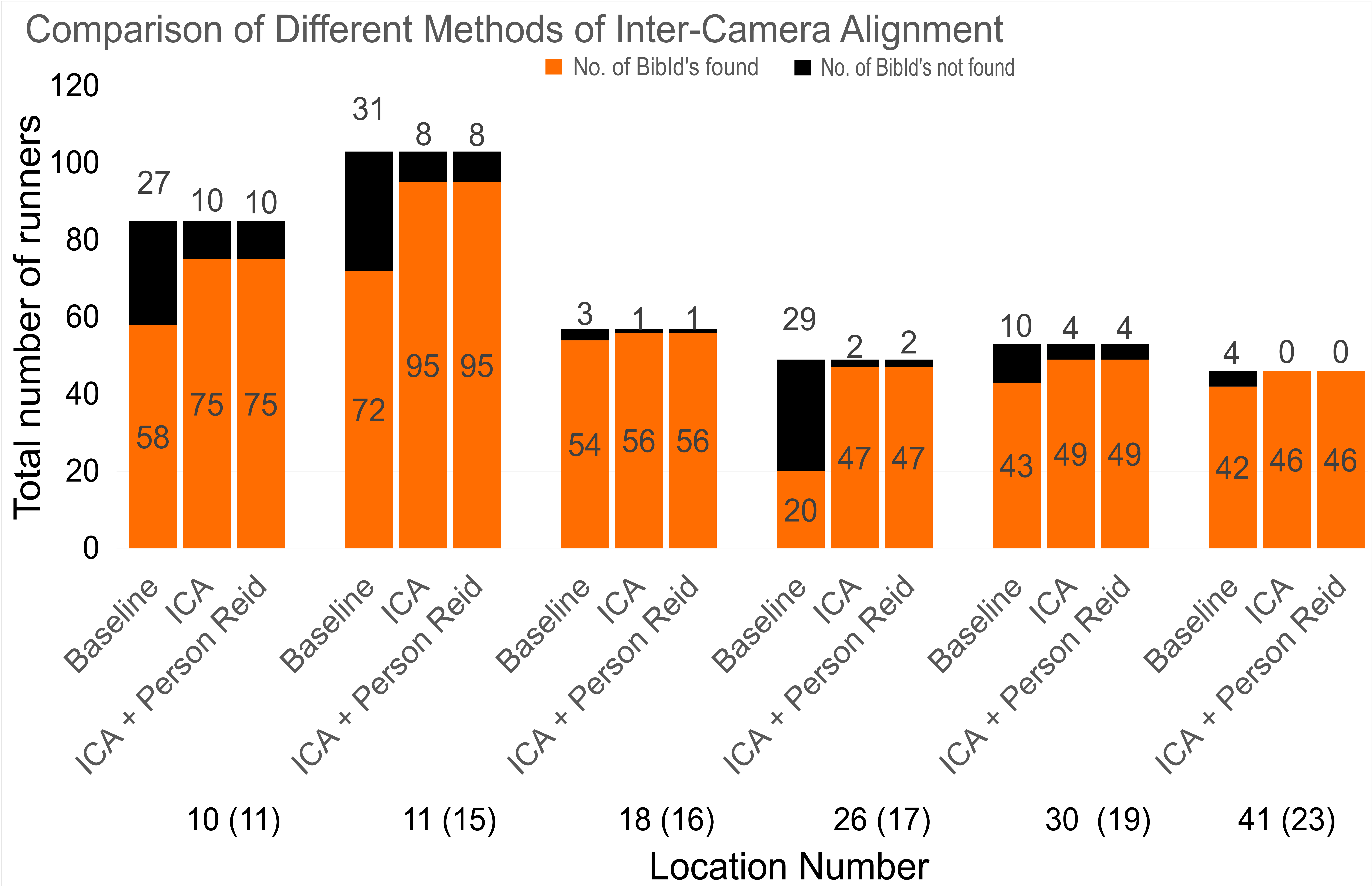} }}%
    \hfill
    \subfloat[Annotation time comparison of different methods of ICA]{{\includegraphics[width=0.48\textwidth]{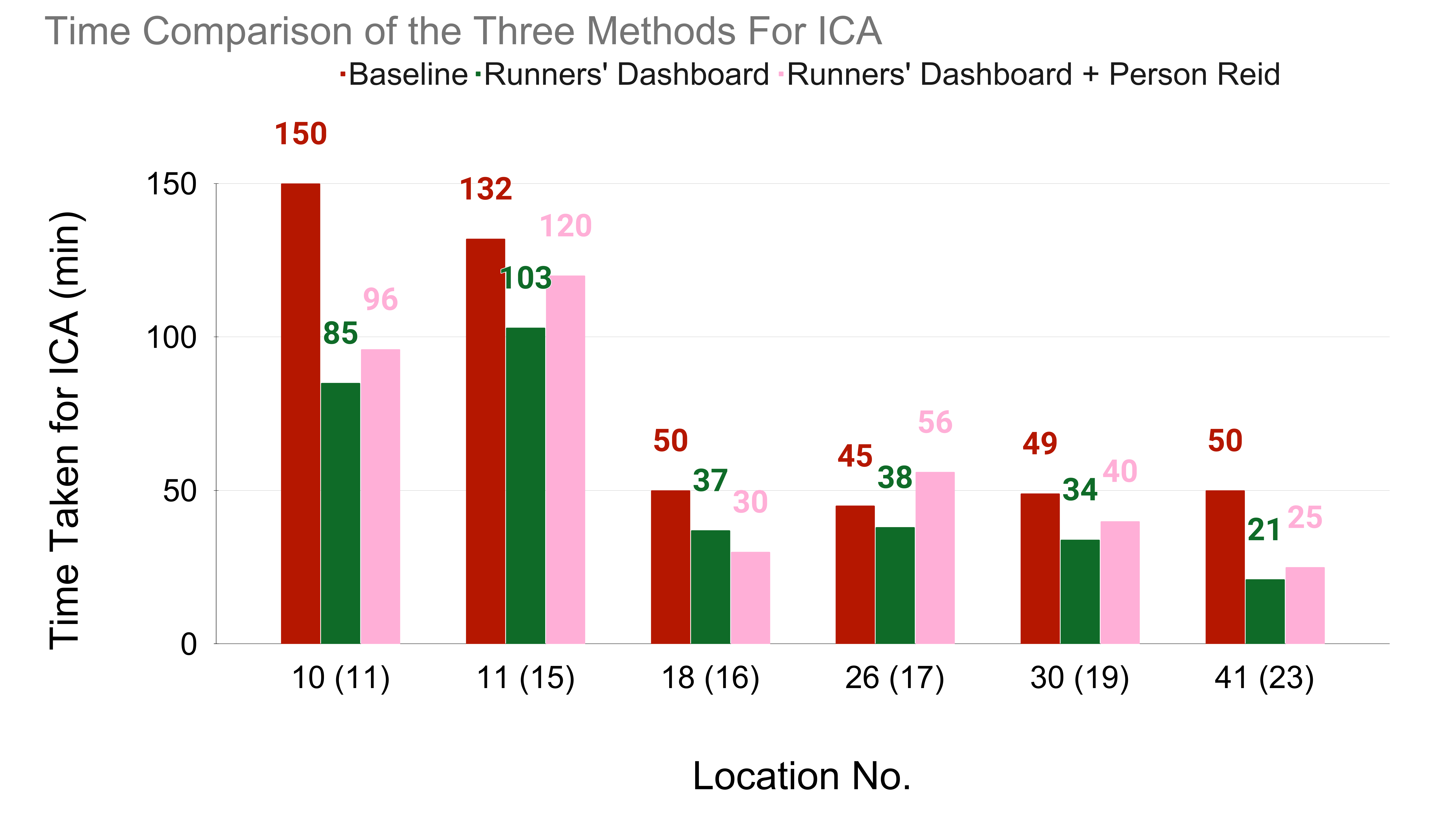} }}%
\end{center}
   \caption{\textbf{Comparing the results of three methods for the Inter-Camera Alignment(ICA) of runners.} The numbers shown in brackets on the horizontal axis are the scores of each location, indicating the difficulty level of the scenarios present in videos, refer to Table \ref{table:scoreDistribution} for more details. (a) The number of runners with bibId not recognizable are higher in case of baseline method. Using ICA tool (runners' dashboard), most of the runners' bibId are identified. A few number of unidentified runners using tool reflects the number of those runners that did not finish the race, and hence were unidentifiable using the tool. (b) It is clear in the figure that ICA using runners' dashboard is the quickest of all the methods. Integration of person re-id in runners' dashboard doesn't decrease the time, however, it is easier and is recommended to use in finding unidentified runners.}
\label{fig:icaResults}
\end{figure*}

\begin{figure}[ht]
    \centering
    \includegraphics[width=0.75\linewidth,]{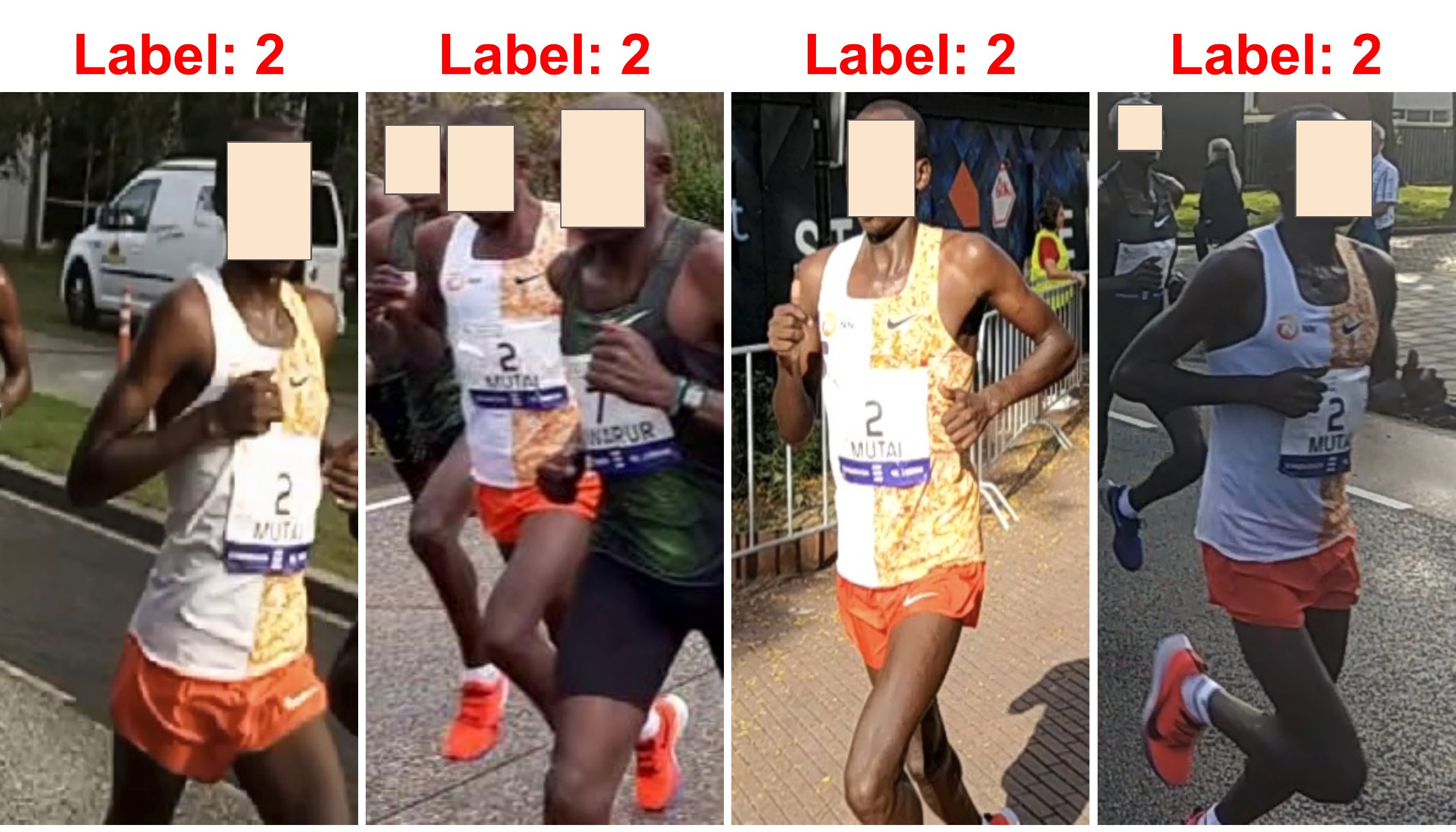}
    \caption{\textbf{An example of inter-camera alignment.} All the appearances of the runner in different non-overlapping cameras should have the same label.}
    \label{fig:cca}
\end{figure}

In inter-camera/cross-camera alignment, we need to ensure that every runner is assigned the same identity across all his appearances in different camera locations. For example, if a runner has the label 'X' in camera location 1, then it should have the same label in all his/her appearances in the remaining camera locations, an example of which is shown in Figure \ref{fig:cca}.

% \begin{figure}[ht]
%     \centering
%     % \subfloat[Sample images scraped from official website]
%     {\includegraphics[width=0.23\linewidth,height=0.4\linewidth]{images/Experiments/ICA/ica1.png} }
%     {\includegraphics[width=0.23\linewidth,height=0.4\linewidth]{images/Experiments/ICA/ica2.png} }
%     {\includegraphics[width=0.23\linewidth,height=0.4\linewidth]{images/Experiments/ICA/ica3.png} }
%     {\includegraphics[width=0.23\linewidth,height=0.4\linewidth]{images/Experiments/ICA/ica4.jpg} }
%     %
%     \caption{An example of inter-camera alignement. All the appearances of the runner should have the same label.}%
%     \label{fig:CCA}%
    
% \end{figure}

Our main objective is to find ways to ensure cross-camera alignment for all the runners, at the cost of minimum time and effort.
There are a number of issues that make the ICA problem difficult. Firstly, there are too many camera locations and runners, which makes the problem more laborious. Secondly, due to crowdedness, occlusion, poor resolution, and illumination in videos, it becomes difficult to see the runners' bibId. In that case, we have to find ways, to search for the identity of the runner in a different camera location. It is tiresome to search a runner amongst thousands of other runners in a pool of videos with hours of recording. Therefore, there's a need for some better way to do the inter-camera alignment of the runners.

\textbf{Setup.}\label{subsec:icaSetup} In all the experiments for the current problem, we have used the sample dataset already discussed in Section \ref{sec:sampleDataset}.
As the main objective of the experiment is to align the identities of the runners in all the 6 cameras of the sample dataset, we are only interested in the bibId of the runners and the frame number in which the runner's bibId is clearly visible. In case, there is no bibId attached on the runner's chest or in case it's taking too long or if it is not possible to find the bibId of the runner, then a unique ID is assigned in the pattern 'LiRj', where 'L' stands for location, 'R' stands for the runner, and 'i' and 'j' represents the location and the runner number respectively.

% \begin{figure*}
%     \centering
%     \includegraphics[width=\linewidth]{images/Experiments/ICA/Time Comparison of the Three Methods For ICA.png}
%     \caption{\textbf{Comparing the results of three methods for the Inter-Camera Alignment(ICA) of runners.} The numbers shown in brackets on the horizontal axis are the scores of each location, indicating the difficulty level of the scenarios present in videos, refer to Table \ref{table:scoreDistribution} for more details. It is clear in the figure that ICA using runners' dashboard is the quickest of all the methods. Integration of person re-id in runners' dashboard doesn't decrease the time, however, it is easier and is recommended to use in finding unidentified runners.}
%     \label{fig:icaResults}
% \end{figure*}

\subsection{Exp 7: Baseline}
In the baseline experiment, the annotator has to find the frame where the runners' bibId is recognizable. In case of a specific runner, if it is not possible to find the bibId due to discrepancies mentioned in Section \ref{subsec:scenarios}, or if the tag of the runner is missing, then the annotator will try to find the runner in videos from a different camera location. The new camera location can be randomly selected from the locations having location number lower than the current location number, as it is assumed that the runner has not evaded any checkpoint and has passed through all the previous locations before reaching the current location. If the annotator can locate the runner in the videos of randomly selected location then he will assign the runners' bibId as it's label, and if he fails to find the runner after spending a few minutes then he will assign it a unique Id as mentioned in Section \ref{subsec:icaSetup}.

\textbf{Analysis.} The results of the experiments are shown in Figure \ref{fig:icaResults}. It can be observed in the Figure that the number of unidentified runners is quite high, if we try to find the runners naively in other video locations. Only 60.71\% runners are identified. It took almost 8 hrs to perform the inter-camera alignment in the sample dataset. Also, many runners are still unidentified. Therefore, a better solution is needed for the inter-camera alignment of the runners.

\subsection{Exp 8: Creating an Inter-Camera Alignment Tool}
\label{subsec:icaExperiment}
To ease the problem of aligning the runners across multiple cameras, we have created a dashboard of runners as mentioned in Section \ref{subsec:icaTool}. Therefore, in this experiment, the task is performed in a similar manner as in the baseline method, except we try to locate an unidentified runner using the ICA tool. To find a runner, we can use a partial name search or bibId search option in the tool. If no part of the name or bibId is visible, then we select a nearby runner whose bibId or name is visible. Firstly, we find the time 't' at which this specific runner reaches the current location. We use this time to find all the runners crossing the current location, in time duration t-$\delta$t $\leqslant$ t $\leqslant$ t+$\delta$t, where $\delta$t = 1 min. This way we reduce the search space, and can quickly identify the runner using images provided in the tool. If the runner's identity is unknown even after spending a few minutes searching it, then assign a unique id to the runner as mentioned in Section \ref{subsec:icaSetup}.

\textbf{Analysis.} The results of the experiment are shown in Figure \ref{fig:icaResults}. It can be noticed that there's a big drop in the overall annotation time when the runners' dashboard tool is used for the inter-camera alignment of runners. The tool is almost 1.5x faster than the naive baseline method. Also, the number of unidentified runners is also very less in comparison to the baseline method. Total 93.64\% runners are identified using the ICA tool. The small number of unidentified runners' can also be explained. The tool comprises the information of runners' who finished the race. It doesn't contain the details of runners that abandoned the race in the middle. That's why it is not possible to find the identity of such runners using the dashboard tool.

Overall, the tool is user friendly and quick in locating unidentified runners. But, it is a bit tricky to use the tool, and hence needs some time to get the expertise. Also, a manual search operation is performed every time to find the runner. So, we will try to find some semi-automatic way to reduce the number of manual search operations.

\subsection{Exp 9: Intelligent use of noise for Inter-Camera Alignment}

To reduce the effort of manually searching the runners in the dashboard tool, we try to automatize the process. To do this, we intelligently use the noisy results of the person re-identification system. Person re-identification is a well-known research problem in computer vision, applications of which ranges from tracking persons appearing across multiple non-overlapping cameras. The main goal of a person re-id is to retrieve all appearances of a person from a large gallery of images captured from cameras with non-overlapping views. There is a lot of work done in the past on supervised and unsupervised person re-identification. Though supervised methods can attain good performance, they need large-scale labeled datasets. 

\begin{figure}[ht]
    \centering
    \includegraphics[width=\linewidth]{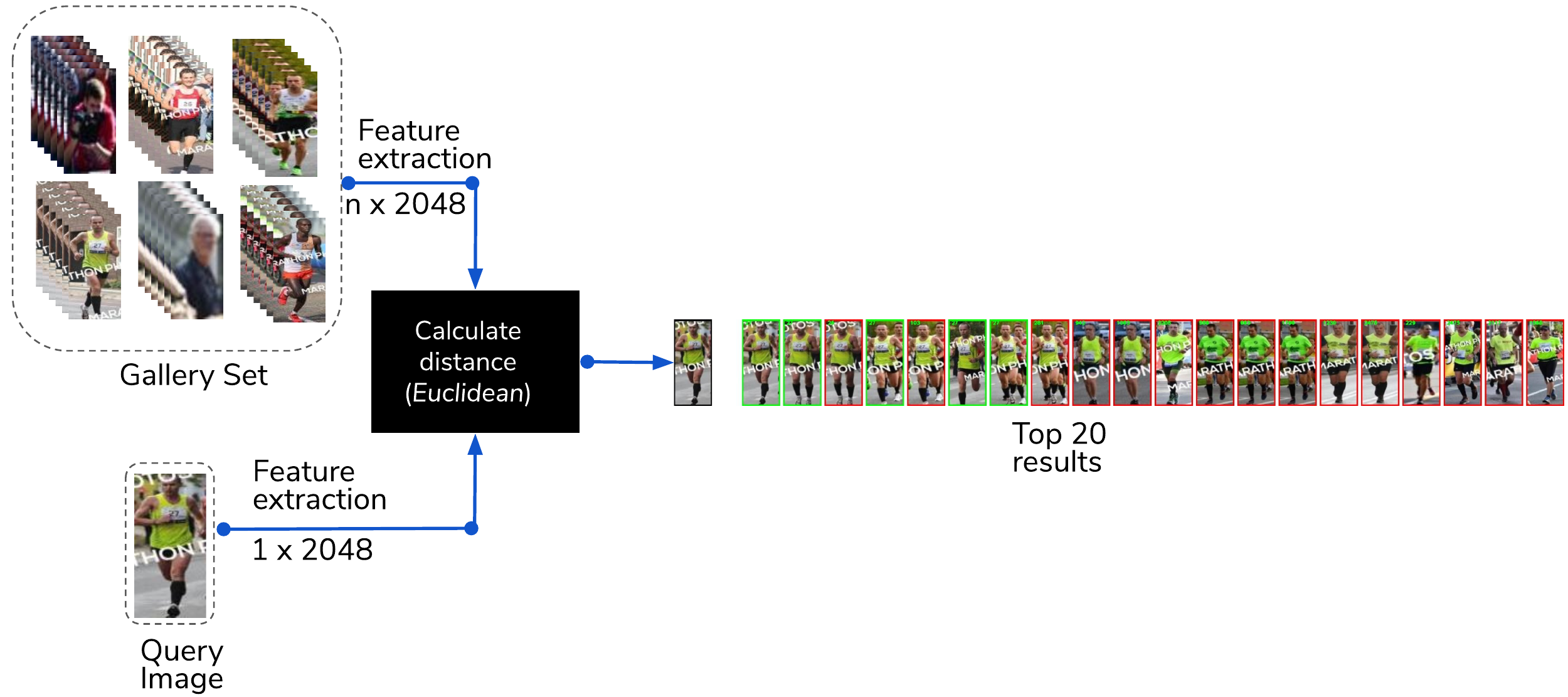}
    \caption{\textbf{Workflow diagram of the person re-id model.} Firstly, the feature vector of the gallery images and the probe image is calculated. Euclidean distance is calculated between the gallery images and probe image feature vectors and the 20 closest images are displayed as the final output.}
    \label{fig:personReidWorkflow}
\end{figure}

\begin{figure}[ht]
\begin{center}
     \subfloat[Gallery browser]{{\includegraphics[width=0.54\linewidth]{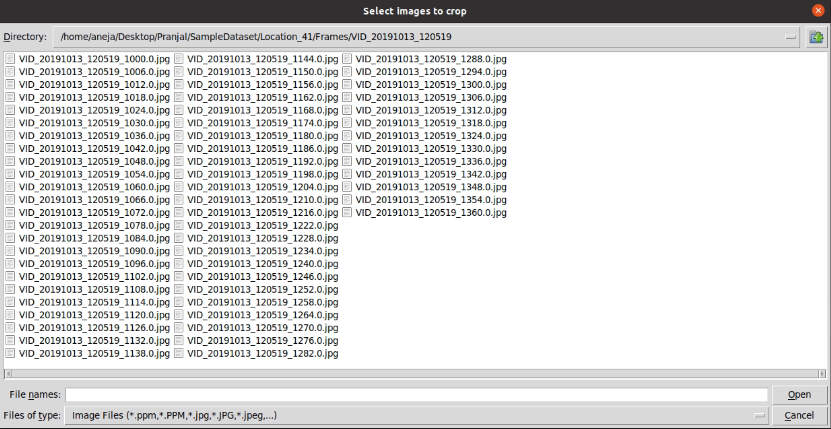} }}%
     \hfill
     \subfloat[Image cropper tool]{{\includegraphics[width=0.42\linewidth]{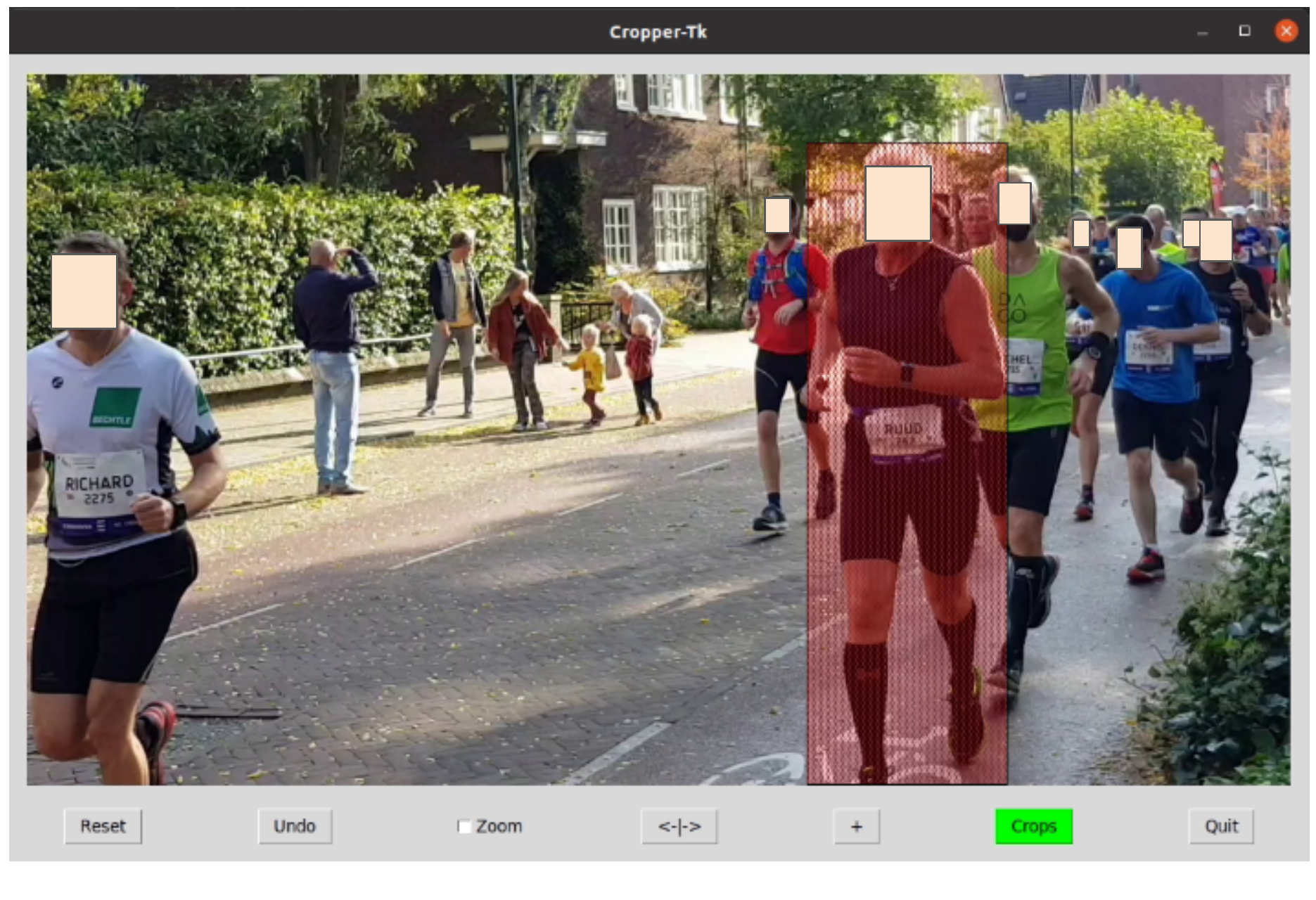} }}%
    \hfill
    \subfloat[Visualization of top 20 results]{{\includegraphics[width=\linewidth]{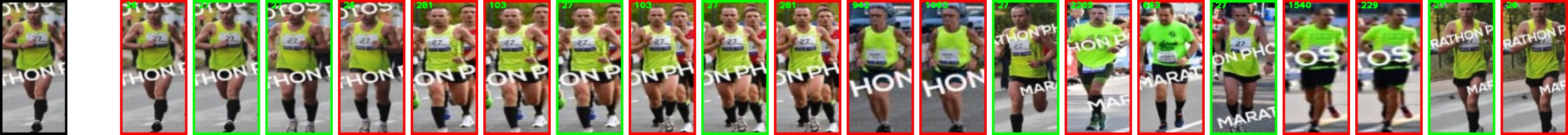} }}%
    
\end{center}
   \caption{\textbf{An outlook of the person re-identification tool integrated with runners' dashboard.} (a) The tool offers the option to quickly browse images from the gallery set. (b) The target runner can be cropped from the selected image. After cropping, the person re-id system will be activated for the cropped person's image. (c) The top-20 matches closest to the probe images are displayed as the final output.}
\label{fig:personReidGUI}
\end{figure}

\begin{figure*}[ht]
    \centering
    \includegraphics[width=\linewidth]{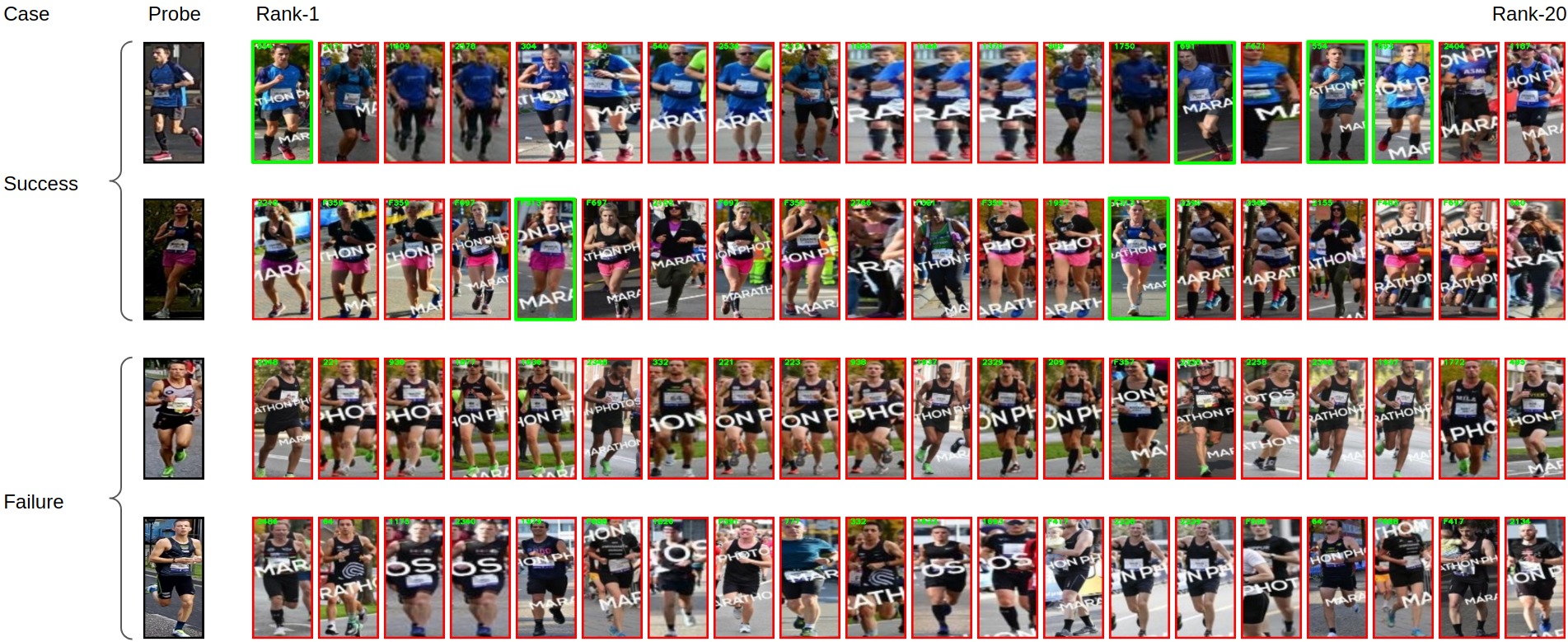}
    \caption{\textbf{Results of the inter-camera alignment experiment using person re-id model.} The upper two rows correspond to the cases where the tool successfully finds at least 1 image of the target runner. The lower two rows correspond to cases where the tool fails to find any image correct image of the target runner in the top-20 results. In a successful case, the matching images of the runner have different labels. It is because the target runner may have appeared in the background of an image of some different runner, and the identity of the other runner is displayed in the result.}
    \label{fig:personReidResults}
\end{figure*}

\textbf{Setup.} In our experiment, we use an off-the-shelf person re-id~\cite{torchreid}, to identify the runner among the gallery of images of all the participant runners. In the person re-id task, we used the dataset of runners, scraped from the official website, as mentioned in Section \ref{dataCollection}. The dataset consists of images with a watermark text in every image. The text isn't removed as it was not affecting the re-id model results. As, we used an object detector~\cite{yolov3}, to crop runners from the images, the dataset also contains random images of non-runners appearing in the background. Hence, the dataset is noisy and contains some garbage images as well. Our main interest is to see if we can use the noisy results of the person re-id model to find the unidentified runner. We used the pre-trained model, trained on market1501~\cite{market1501}. In~\cite{torchreid}, the dataset is divided into a gallery set and a query set. In our experiment, the gallery set consists of at-most 20 images per runner, and minimum 2 images per runner, whereas the query set consists of images, used as probe images to find all the appearances of the runner in the gallery set.

A person re-id tool is developed and integrated into the runners' dashboard. An outlook of the tool is shown in Figure \ref{fig:personReidGUI}. The tool helps in quickly browsing the images from the gallery. It also allows the user to crop the person of interest from the image. The tool takes the probe image of the runner as input and process out the top 20 images closest to the probe image. Firstly, the feature vector of the probe image and all the images in the gallery set is created. Then, the euclidean distance between the feature vector of the probe image with the feature vector of all the gallery images is calculated. The top-20 images having the minimum distance with the probe image are selected as output. In the end, top-20 results are displayed as shown in the workflow diagram in Figure \ref{fig:personReidWorkflow}.

% \vspace{-0.105cm}

In the experiment, the same steps are followed as in the previous Experiment \ref{subsec:icaExperiment}, except that for any unidentified runner whose bibId or name is not visible, instead of finding the runner using the dashboard, we first find the runner using the person re-id tool integrated into runners' dashboard. The runners' dashboard is only used when the runner's name or bibId is visible, or the person re-id tool fails to find the runner.

\textbf{Analysis.} The results are mentioned in the Figure \ref{fig:icaResults}. The method takes very little time in comparison to the baseline method. However, it is observed that the integration of person re-id is taking more time than solely using the runners' dashboard tool. The reason is that the person re-id takes around 10-15 seconds per query image, to process the results. As the deep learning models' performances are still not saturated, they tend to give false results, as can be seen in Figure \ref{fig:personReidResults}. Due to which, we had to perform multiple runs in the person re-id to reach the correct result, and in case re-id fails, we had to use the runners' dashboard at the end to find the identity of the runner. As a result, the overall time of annotation is increased. However, it is more flexible and easy to identify runners using person re-id, so we propose to use this method for the inter-camera alignment task.

%------------------------------------------------------------------------

\section{Conclusion}
In this work, we proposed the heuristics to annotate a large-scale in-the-wild video dataset of marathon runners. We discussed the problems that arise in the annotation of the marathon dataset covering real-world scenarios. We demonstrated how to reduce the overall cost of annotation by reducing the frame extraction rate. Additionally, we investigated different ways to generate efficient tight bounding boxes. Our study shows that using box interpolation is the most effective way of generating bounding boxes in such datasets. We also proposed a novel method of aligning the runners in the cross-camera setting of multiple non-overlapping cameras. We introduced an inter-camera alignment tool integrated with state-of-the-art deep learning person re-identification method, to help in quickly and efficiently aligning unidentified runners across multiple disjoint cameras.

Even though our methods significantly reduce the human effort and total cost of annotation, more research into cross-camera alignment can incentivize the annotation time and accuracy. In this paper, we used a noisy dataset for the person re-identification task that contained images of non-runners. Due to this, the person re-id system is not very accurate. The dataset can be more refined and augmented to improve the accuracy. Also, for bounding box regression, more alternatives can be explored in the future. One alternative could be the combination of extreme clicking~\cite{gygli2019efficient}, and box interpolation, to further reduce the annotation time. In future, we will look forward to investigative these incentives.

% You must include your signed IEEE copyright release form when you submit
% your finished paper. We MUST have this form before your paper can be
% published in the proceedings.

% Please direct any questions to the production editor in charge of these 
% proceedings at the IEEE Computer Society Press: 
% \url{https://www.computer.org/about/contact}. 

%------------------------------------------------------------------------

% \section{Future Work}

% \clearpage

{\small
\bibliographystyle{ieee_fullname}
\bibliography{egbib}
}

\clearpage

\appendix

\section{Kolmogorov-Smirnov (KS) Test}
\label{apdx:ksTest}

For larger sample sizes, the approximate critical value $D_{\alpha}$ is given by the equation,\newline

\begin{equation}
    D_{\alpha } = c{(\alpha)}\sqrt{\frac{n_{1} + n_{2}}{n_{1}n_{2}}}
\end{equation}

where, $n_{1}$ and $n_{2}$ are the sample sizes of the two distributions and $\alpha$ and $c{(\alpha)}$ are the coefficients given by table mentioned ~\cite{kstest}.\newline

Original dataset location scores  from the table \ref{table:originalDatasetScores} = $S_{originalDataset}$ = \{8, 10, 11, 12, 13, 14, 15, 15, 15, 16, 16, 16, 16, 16, 16, 17, 17, 17, 17, 18, 18, 18, 18, 18, 19, 19, 20, 20, 20, 21, 22, 22, 23, 23, 24\}\newline

Sample size of $S_{originalDataset}$ = 35\newline

As we want to select 6 scores for the sample dataset,\newline

Sample size of $S_{sampledDataset}$ = 6,\newline

$c{(\alpha)}$ = 1.63, $\alpha$ = 0.01 \newline

$D_{\alpha } = 1.63\sqrt{\frac{35 + 6}{35\times 6}} = 0.7202$.\newline

The KS test gives statistics and $p_{value}$ as output.\newline

Null hypothesis is that the two distributions are different.\newline

\begin{equation}
    \text{if}, 
    \begin{cases}
    \text{statistics} < D_{\alpha} \\ \text{and p-value} > {\alpha}, & \textit{null hypothesis rejected}\\\\
    \text{else},  &  \textit{null hypothesis accepted.}\\
    \end{cases}
\label{eq:IoU}
\end{equation}

% if $\left \{  \right. statistics < D_{\alpha} $  and  $p_{value} > {\alpha}$, then null hypothesis rejected\newline

% else, $\left \{  \right.$ hypothesis accepted\newline

After thousands of iterations of random sampling the score values, we selected 6 best scores that have the distribution similar to the original dataset score distribution.\newline

Sampled dataset location scores = $S_{sampled_Dataset}$ = \{11, 15, 16, 17, 19, 23\}

\section{Webscraping}
\label{apdx:webscraping}

\begin{table}[ht]
\begin{center}
\begin{tabular}{  p{2cm}  p{5cm} } 
\toprule
\textbf{Data Type} & \textbf{Field name}\\ 
\midrule
Full marathon data & id, eventId, raceId, bib, bibForUrl, category, rank, genderRank, categoryRank, gunTime, chipTime primaryDisplayTime, speedInKmh, name, countryCode, activityType, gender, city, cumulativeTime\_5k, name\_5k, cumulativeTime\_10k, name\_10k, cumulativeTime\_15k, name\_15k, cumulativeTime\_20k, name\_20k, cumulativeTime\_25k, name\_25k, cumulativeTime\_half, name\_half, cumulativeTime\_30k, name\_30k, cumulativeTime\_35k, name\_35k, cumulativeTime\_40k, name\_40k, cumulativeTime\_finish, name\_finish, gunTimeInSec, chipTimeInSec, customValues, displayDistance, qualified\\ 
\addlinespace
Half marathon data & id, eventId, raceId, bib, bibForUrl, category, rank, genderRank, categoryRank, gunTime, chipTime primaryDisplayTime, speedInKmh, name, countryCode, activityType, gender, city, cumulativeTime\_5k, name\_5k, cumulativeTime\_10k, name\_10k, cumulativeTime\_15k, cumulativeTime\_finish, name\_finish, gunTimeInSec, chipTimeInSec, customValues, displayDistance, qualified \\ 
\bottomrule
\end{tabular}
\end{center}
\caption{\textbf{Field names corresponding to which the data of full-marathon and half-marathon is scraped from the official website.}}
\label{table:fieldNames}
\end{table}

For scraping the data, python libraries namely Selenium web-driver~\cite{selenium} and Beautiful-soup~\cite{beautifulsoup} are used. The data is available on the website of Eindhoven Marathon~\cite{officialWebsite}. The request was first intercepted using the selenium web-driver and the data available in JSON format on the website was retrieved corresponding to all the page requests containing the data. Only data corresponding to the required fields is read and later saved into a \textit{xls} file.

\begin{table*}[t]
\begin{center}
\begin{tabular}{ p{2cm}  p{2cm}  p{2cm}  p{2cm}  p{2cm}  p{2cm}  p{1cm} }
\toprule
        \textbf{Location} & \textbf{Occlusion} & \textbf{Lighting} & \textbf{Recording Angle} & \textbf{Resolution} & \textbf{\#Crowded Videos} & \textbf{Score}\\\midrule
        5 & 1 & 3 & 2 & 1 & 1 & 8\\
        3 & 1 & 3 & 2 & 3 & 1 & 10\\
        10 & 3 & 1 & 1 & 4 & 2 & 11\\
        2 & 1 & 2 & 4 & 4 & 1 & 12\\
        4 & 1 & 4 & 4 & 3 & 1 & 13\\
        9 & 2 & 3 & 3 & 4 & 2 & 14\\
        25 & 2 & 4 & 2 & 5 & 2 & 15\\
        11 & 2 & 3 & 5 & 2 & 3 & 15\\
        35 & 3 & 3 & 5 & 2 & 2 & 15\\
        8 & 2 & 3 & 5 & 4 & 2 & 16\\
        28 & 5 & 3 & 3 & 2 & 3 & 16\\
        29 & 3 & 4 & 3 & 3 & 3 & 16\\
        33 & 4 & 4 & 2 & 2 & 4 & 16\\
        22 & 3 & 2 & 5 & 4 & 2 & 16\\
        30 & 3 & 3 & 5 & 3 & 2 & 16\\
        37 & 4 & 4 & 4 & 2 & 3 & 17\\
        40 & 4 & 4 & 4 & 2 & 3 & 17\\
        26 & 3 & 4 & 4 & 3 & 3 & 17\\
        14 & 2 & 4 & 5 & 4 & 2 & 17\\
        12 & 3 & 5 & 3 & 4 & 3 & 18\\
        13 & 3 & 4 & 4 & 3 & 4 & 18\\
        31 & 4 & 4 & 4 & 3 & 3 & 18\\
        14 & 3 & 5 & 4 & 4 & 2 & 18\\
        17 & 3 & 4 & 5 & 4 & 2 & 18\\
        20 & 4 & 4 & 3 & 4 & 4 & 19\\
        18 & 4 & 3 & 5 & 4 & 3 & 19\\
        15 & 4 & 4 & 5 & 3 & 4 & 20\\
        38 & 4 & 4 & 5 & 4 & 3 & 20\\
        39 & 4 & 4 & 5 & 3 & 4 & 20\\
        36 & 4 & 4 & 5 & 4 & 4 & 21\\
        34 & 4 & 4 & 5 & 5 & 4 & 22\\
        42 & 4 & 5 & 5 & 5 & 3 & 22\\
        27 & 5 & 5 & 5 & 4 & 4 & 23\\
        41 & 4 & 5 & 5 & 5 & 4 & 23\\
        24 & 5 & 5 & 5 & 4 & 5 & 24\\

\bottomrule
\end{tabular}
\end{center}
\caption{\textbf{Score distribution of recording locations in original dataset.} Lower score means bad recording scenarios and higher score means good recording scenarios.}
\label{table:originalDatasetScores}
\end{table*}

\end{document}